\documentclass[10pt,twocolumn,letterpaper]{article}

\usepackage{cvpr}              
\usepackage{multirow}
\usepackage{pifont}
\usepackage[linesnumbered,ruled]{algorithm2e}
\usepackage[table]{xcolor}

\newcommand\blfootnote[1]{%
\begingroup
\renewcommand\thefootnote{}\footnote{#1}%
\addtocounter{footnote}{-1}%
\endgroup
}

\definecolor{cvprblue}{rgb}{0.21,0.49,0.74}

\definecolor{pastelyellow}{RGB}{255, 255, 102}
\definecolor{pastelred}{RGB}{255, 105, 97}
\definecolor{pastelgreen}{RGB}{102, 255, 102}
\definecolor{pastelblue}{RGB}{102, 178, 255}
\definecolor{oceanicblue}{RGB}{0, 119, 190}
\definecolor{oceanicgreen}{RGB}{0, 166, 147}
\usepackage[pagebackref,breaklinks,colorlinks,citecolor=cvprblue]{hyperref}

\def\paperID{15729} 
\def\confName{CVPR}
\def\confYear{2024}

\title{Visual Foundation Models Boost Cross-Modal Unsupervised Domain Adaptation for 3D Semantic Segmentation}

\author{Jingyi Xu$^1$, Weidong Yang$^{1,\dagger}$, Lingdong Kong$^2$, Youquan Liu$^3$, Rui Zhang$^1$, Qingyuan Zhou$^1$, Ben Fei$^{1,\dagger}$\\
$^1$ Fudan University, $^2$ National University of Singapore, $^3$ Hochschule Bremerhaven\\
{\tt\small jyxu22@m.fudan.edu.cn, wdyang@fudan.edu.cn, bfei21@m.fudan.edu.cn}
}

\begin{document}


\twocolumn[{%
\renewcommand\twocolumn[1][]{#1}%
\maketitle
\begin{center}
    \centering
    \captionsetup{type=figure}
    \includegraphics[width=\textwidth]{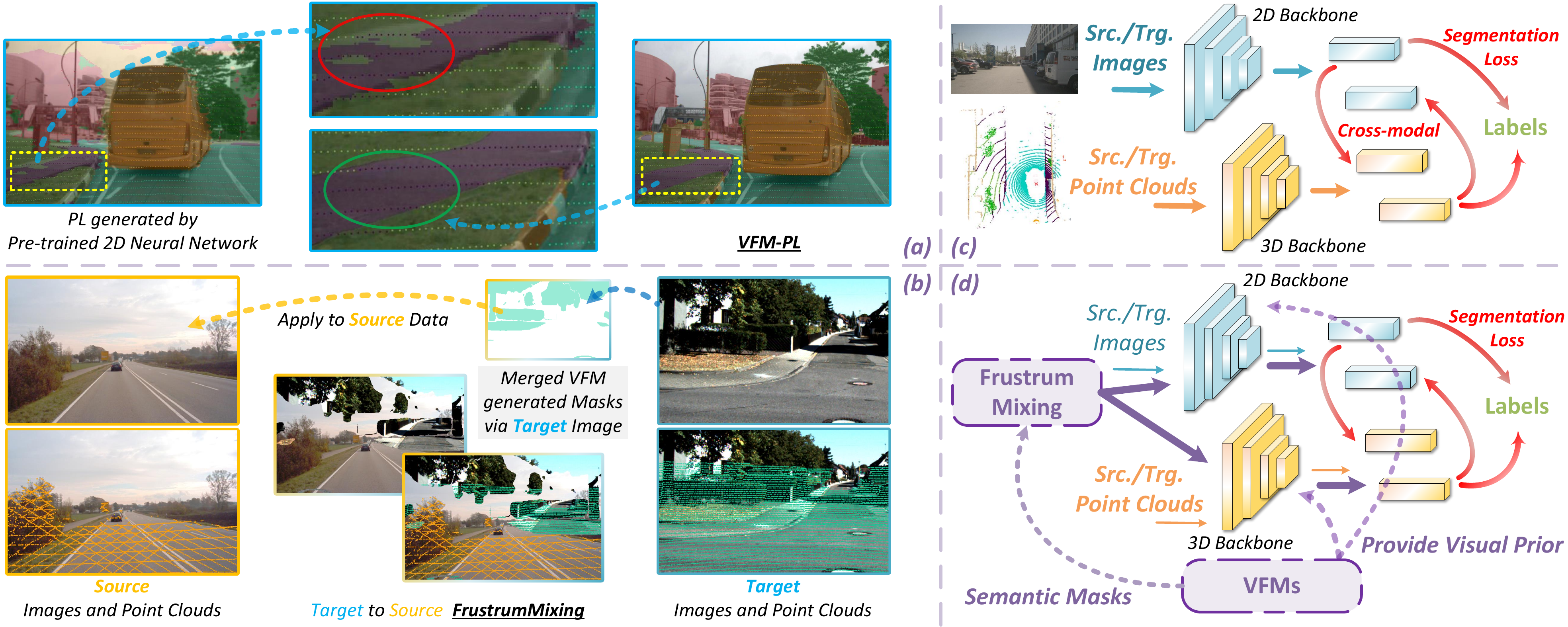}
    \vspace{-16pt}
    \captionof{figure}{(a). Comparison between the traditional pseudo labels (\textbf{Left}) and pseudo labels from our \textbf{VFM-PL} (\textbf{right}). (b). Illustration of our \textbf{FrustumMixing}, narrowing the domain gap by mixing the source and target samples with the help of VFMs. Comparison of (c) traditional cross-domain UDA methods and (d) \textbf{VFMSeg}, where VFMSeg leverages the powerful prior of VFMs to boost UDA performance.}
    \label{teaser}
\end{center}%
}]

\blfootnote{{$\dagger$}Corresponding author}

\begin{abstract}
Unsupervised domain adaptation (UDA) is vital for alleviating the workload of labeling 3D point cloud data and mitigating the absence of labels when facing a newly defined domain.
Various methods of utilizing images to enhance the performance of cross-domain 3D segmentation have recently emerged. 
However, the pseudo labels, which are generated from models trained on the source domain and provide additional supervised signals for the unseen domain, are inadequate when utilized for 3D segmentation due to their inherent noisiness and consequently restrict the accuracy of neural networks.
With the advent of 2D visual foundation models (VFMs) and their abundant knowledge prior, we propose a novel pipeline \textbf{VFMSeg} to further enhance the cross-modal unsupervised domain adaptation framework by leveraging these models. In this work, we study how to harness the knowledge priors learned by VFMs to produce more accurate labels for unlabeled target domains and improve overall performance. 
We first utilize a multi-modal VFM, which is pre-trained on large scale image-text pairs, to provide supervised labels (\textbf{VFM-PL}) for images and point clouds from the target domain. 
Then, another VFM trained on fine-grained 2D masks is adopted to guide the generation of semantically augmented images and point clouds to enhance the performance of neural networks, which mix the data from source and target domains like view frustums (\textbf{FrustumMixing}).
Finally, we merge class-wise prediction across modalities to produce more accurate annotations for unlabeled target domains. 
Our method is evaluated on various autonomous driving datasets and the results demonstrate a significant improvement for 3D segmentation task.
Our code is available at \href{https://github.com/EtronTech/VFMSeg}{https://github.com/EtronTech/VFMSeg}

\end{abstract}    
\section{Introduction}
\label{sec:intro}
Point cloud segmentation is vital for real-world applications such as 3D scene perception, robotics, and autonomous driving~\cite{wei2020multi, sanchez2023domain}.
During this process, each individual point within the point cloud is assigned a semantic label to enhance understanding and analysis~\cite{shin2022mm, zhao2021few}.
However, labeling massive 3D data is a laborious and costly process~\cite{hu2021towards, zhang2023growsp}. 
Hence, it is significant to develop \textbf{domain adaptation}, i.e. \textbf{unsupervised}, methods that could efficiently exploit existing annotation of the source point cloud and transfer the acquired knowledge to the label-free target 3D domain~\cite{jiang2021guided}.
Otherwise, assigning semantic labels for point clouds is intrinsically challenging due to their sparsely distributed, unstructured, and colorless nature~\cite{hu2022sqn}. 
To incorporate multi-modal information, the advent of multi-modal autonomous driving datasets~\cite{behley2019semantickitti,caesar2020nuscenes,gaidon2016virtual,feng2021dataset} has facilitated the availability of concurrent images alongside point clouds, which opens a valuable research topic and enables researchers to utilize the rich visual information embedded in images that vastly facilitate 3D semantic segmentation. 

Recent research has proposed a promising line of frameworks~\cite{jaritz2020xmuda,liu2021adversarial,peng2021sparse,cardace2023exploiting} that simultaneously leverage multi-modal to address the 3D segmentation task in the unlabeled target domain. 
In these approaches, the neural network for different modalities was first pre-trained and then applied to the target domain for generating pseudo-labels (PL)~\cite{li2019bidirectional}. 
The utilization of these pseudo-labels in the subsequent training stage can provide supervision for the target domain, thereby improving the overall performance.
Despite the proven effectiveness of this method, pseudo-labels are inevitably noisy (Fig.~\ref{teaser}a \textbf{Left}).
These noises arise from the limited capacity of pre-trained models, which consequently restricts the segmentation capability of neural networks.
Besides, the information exchange across various domains is realized through shared neural networks, resulting in limited adaptation at a coarse level~\cite{yang2023label}.

Visual Foundation Models (VFMs) have already demonstrated remarkable performance on a variety of open-world 2D vision tasks~\cite{Alexander2023sam,zou2023generalized,zou2023seem,wang2023hierarchical,wang2023seggpt}.
Specifically, the Segment Anything Model (SAM)~\cite{Alexander2023sam} has achieved outstanding performance on zero-shot 2D segmentation, while Segment Everything Everywhere Model (SEEM)~\cite{zou2023seem} further extends such capability of SAM by providing accurate semantic labels for generated masks. 

In the light of rich and robust visual priors learned by VFMs~\cite{rozenberszki2022language}, we propose \textbf{VFMSeg} to fully exploit their zero-shot segmentation capability of VFMs and transfer 2D visual knowledge across modalities and domains.
To tackle the inaccurate PLs generated by pre-trained models, we present \textbf{VFM-PL}, which generates more precise pseudo-labels (Fig.~\ref{teaser}a \textbf{Right}) by taming SEEM for labeling images from autonomous driving datasets.
Furthermore, aiming to further narrow the gap between the source and target domain, we have additionally developed a method dubbed \textbf{FrustumMixing}, as illustrated in Fig.~\ref{teaser}b.
FrustumMixing utilizes SAM~\cite{Alexander2023sam} to generate fine-grained masks for images from both domains. 
These masks are then utilized to mix cross-modal and cross-domain samples, with a portion of the masks involved in the mixing process.
Since the masks generated by SAM lack semantic meaning, we adopt SEEM to complement the absence of semantic labels.
FrustumMixing operates similarly to the concept of view frustum and excels in generating semantically augmented images and point clouds by combining different perspectives.
The inclusion of these semantically augmented samples, which encompass fine-grained semantic instances extracted from the other domain, is anticipated to provide significant performance improvement when feeding into neural networks~\cite{qiu2021semantic}.
To assess the effectiveness of our proposed method, we conducted comprehensive experiments in various cross-modal 3D UDA segmentation scenarios. Our results demonstrate that our method significantly outperforms existing off-the-shelf approaches by a substantial margin.


Our contributions of the proposed VFMSeg are fourfold: 
\begin{itemize}
    \item We propose VFMSeg, a novel cross-modal unsupervised domain adaptation framework that boosts the performance of 3D semantic segmentation by the merit of visual foundation models.
    \item To tackle the inaccuracy of traditional pseudo labels, we exploit the knowledge priors learned by VFMs to produce more precise labels for target domain. 
    \item To further narrow the domain gap, we leverage another VFM trained on fine-grained 2D masks to guide the generation of semantically augmented images and point clouds, thereby enhancing the cross-domain capability of the backbones.
    \item Extensive experiments on three cross-domain settings demonstrate our VFMSeg can outperform existing state-of-the-art counterparts.
\end{itemize}

\section{Related Works}

\begin{figure*}[ht]
	\centering
	\includegraphics[width=1\linewidth]{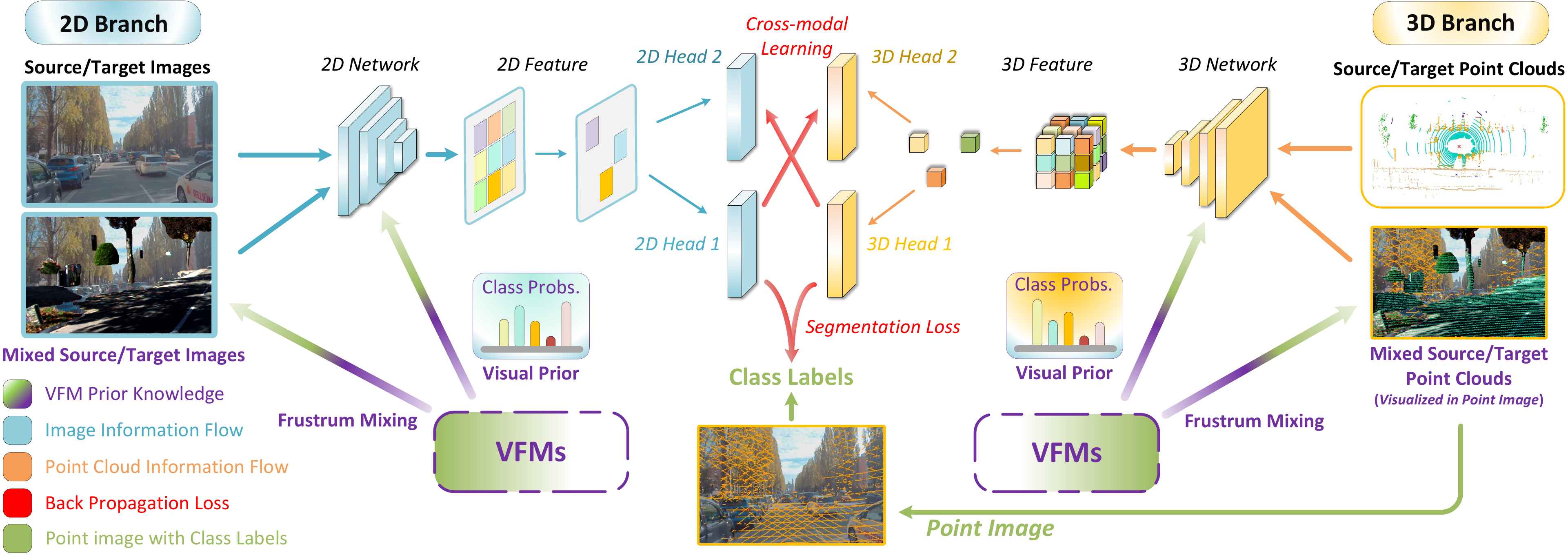}
	\caption{\textbf{Framework overview.} Both 2D and 3D neural networks are trained on source \textit{\textbf{and}} target data. Hence, the domain-invariant feature is captured during parameter optimization. There are two projection heads in those networks. The first head leverages supervision signal within labels and the second head provides cross-modal information exchange through KL-Divergence (Sec \ref{subsec:framework}). 
    Since the target domain is free of labels under the UDA setting, pre-trained 2D and 3D networks are first utilized to generate pseudo-labels for the target domain.
    VFM is applied to provide guidance for producing more accurate pseudo-labels (Sec \ref{subsec:leverageVFM}).
    The visual prior of a VFM is also leveraged to create diverse training samples that bridge the gap between two domains (\ref{subsec:frustrumMix}).}
	\label{fig:framework}
\end{figure*}

\textbf{Unsupervised Domain Adaptation for 3D Segmentation.} 
Domain adaptation aims to transfer knowledge and bridge the distribution gap between source and target domains \cite{toldo2020uda}.
For UDA, the source domain has annotations while the target domain is unlabeled and numerous methods have already been proposed to tackle 2D segmentation task \cite{vu2019advent,vu2019dada,pan2020unsupervised,kim2020learning,zhang2020joint}.
UDA for 3D segmentation has drawn great attention in recent studies due to its paramount importance for autonomous driving vehicles~\cite{yi2021dalidar, saltori2022cosmix, rochan2022unsupervised, rochan2022unsupervised}.
Although these methods are promising for uni-modal (image \textbf{or} point cloud) segmentation, the benefit of leveraging complementary information from both modalities has not been fully exploited. 
As a pioneer work, Jaritz et al.~\cite{jaritz2020xmuda,jaritz2022cross} proposed xMUDA framework to capitalize both 2D and 3D modalities for UDA in 3D segmentation. 
Based on that effective framework, Liu et al.~\cite{liu2021adversarial} further incorporate an adversarial training scheme to enhance the information transfer between images and point clouds. 
Peng et al.~\cite{peng2021sparse} introduce a deformable 2D feature patch for better information exchange with 3D point clouds which eventually leads to sufficient domain adaptation. 
Cardace et al.~\cite{cardace2023exploiting} exploit additional depth information to train a 2D encoder that is resistant to domain shift. 
Chen et al.~\cite{chen2023cross} explore a new setting (different from xMUDA) of UDA where the source point clouds are removed from the training process and leverage a mixing strategy for data augmentation to compensate for the absence of source 3D data. 
In this paper, we focus on utilizing VFMs to provide refined supervision in cross-modal UDA.

\noindent\textbf{Visual Foundation Models.} Pre-trained language foundation models~\cite{jia2021scaling,hoffmann2022training,touvron2302llama,waisberg2023gpt} have not only achieved significant advancements in natural language processing (NLP) but also transformed the way people work and conducting research within the community. 
Following this trend, several visual foundation models (VFMs)~\cite{radford2021learning,zou2023generalized,Alexander2023sam,wang2023hierarchical,wang2023seggpt,zou2023seem}  have emerged and showcased their revolutionary capabilities in the field of 2D vision.  Notable VFMs include Segment Anything Model (\textbf{SAM}) \cite{Alexander2023sam}, X-Decoder \cite{zou2023generalized}, Segment Everything Everywhere all at once (\textbf{SEEM}) \cite{zou2023seem}, HIPE \cite{wang2023hierarchical} and SegGPT \cite{wang2023seggpt}. 
These VFMs have made significant contributions to image segmentation tasks and have shown promising potential.
Most recently, VFMs are utilized for various 3D tasks~\cite{liu2023segment, yang2023sam3d, yu20233d}.
However, the fruitful knowledge inherent in these VFMs has not been fully exploited under the UDA 3D segmentation.

\noindent\textbf{Data Augmentation via mixing.}
Deep neural networks commonly exhibit undesirable behaviors, including memorization and overfitting. To address this problem~\cite{yun2019cutmix,zhang2017mixup}, mixing strategies are employed to train neural networks using additional data generated through the convex combination of paired samples and labels. This involves mixing either the entire samples~\cite{zhang2017mixup} or cutting and pasting patches from different samples~\cite{yun2019cutmix}.
Mixing strategies have also demonstrated their effectiveness in mitigating domain shifts in UDA for tasks such as image classification~\cite{wu2020dual,xu2020adversarial} and semantic segmentation~\cite{gao2021dsp, yang2020fda}.
Zou et al.~\cite{zou2021geometry} introduce the concept of Mix3D~\cite{nekrasov2021mix3d} as a pretext task for classification, where the rotation angle of mixed pairs is predicted. 
Kong et al.~\cite{kong2023lasermix} presented a semi-supervised learning pipeline by incorporating a novel LiDAR mixing technique called LaserMix, which intertwines laser beams from different scans to leverage the distinctive spatial prior in LiDAR scenes.
Compositional Semantic Mix (CoSMix)~\cite{saltori2022cosmix} is proposed as the first single-modal UDA approach~\cite{xiao2022polarmix, saltori2023compositional} for point cloud segmentation based on sample mixing.
However, the application of mixing strategies to tackle cross-modal UDA in 3D semantic segmentation has not been fully explored in prior research.
To bridge this research gap, we propose a novel VFM-guided mixing strategy that surpasses the conventional approach of simply concatenating two point clouds or randomly selecting crops. Our VFM-PL takes advantage of VFM to semantically guide the mixing process, thereby enhancing the effectiveness of the mixing strategy.

\section{Method}\label{sec:Method}

\begin{figure}[t]
	\centering
	\includegraphics[width=1\linewidth]{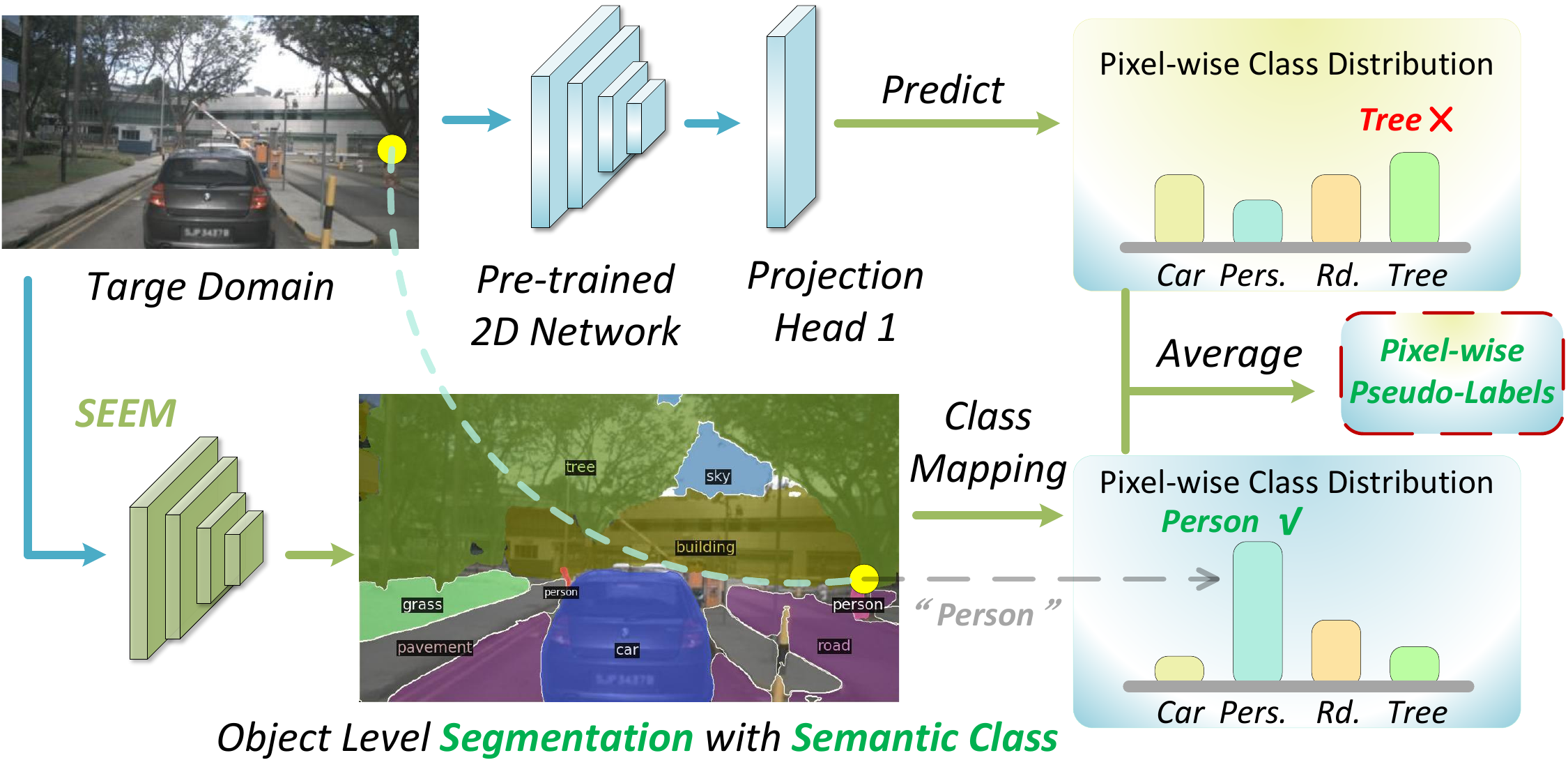}
	
	\caption{\textbf{VFM-PL: Leveraging the visual prior for generating pseudo labels.} We utilize VFM to provide guidance for generating pseudo-labels in the target domain. Since SEEM~\cite{zou2023seem} is trained on a huge amount of image-text pairs and segmentation masks across diverse scenes, its learned feature encoder is naturally resistant to domain shifts. By averaging the probabilistic prediction of pre-trained 2D network and SEEM, the generation of pseudo-labels can be more precise and robust.}
	\label{fig:seem_pl}
\end{figure}

In this section, we first present the overall pipeline of our VFMSeg for cross-modal UDA that leverages both 2D and 3D modalities (Sec \ref{subsec:framework}). Then we elaborate on our proposed VFM-PL of transferring visual prior learned by VFM to source and target domains (Sec \ref{subsec:leverageVFM}). Finally, we introduce the proposed FrustumMixing strategy that further narrows down the domain gap (Sec \ref{subsec:frustrumMix}). 

\subsection{Framework Overview}\label{subsec:framework}

The overall architecture is depicted in Fig.~\ref{fig:framework}. The main steps of the framework can be summarized as follows. 
Initially, we generate the semantically augmented data domain $\boldsymbol{\mathcal{M}}$ by mixing samples from source and target domain with our FrustumMixing (see Sec.~\ref{subsec:frustrumMix}).
Then, we input the data of the source domain $\boldsymbol{\mathcal{S}}$, target domain $\boldsymbol{\mathcal{T}}$, and the mixed source and target domain $\boldsymbol{\mathcal{M}}$ into the 2D and 3D networks. 
This process generates the corresponding feature maps before the classifier, namely $\boldsymbol{\mathcal{F}}_{2\mathrm{D}}^{ \boldsymbol{\mathcal{S}}}$, $\boldsymbol{\mathcal{F}}_{3\mathrm{D}}^{ \boldsymbol{\mathcal{S}}}$, $\boldsymbol{\mathcal{F}}_{2 \mathrm{D}}^{\boldsymbol{\mathcal{T}}}$, and $\boldsymbol{\mathcal{F}}_{3 \mathrm{D}}^{ \boldsymbol{\mathcal{T}}}$.
Following that, our VFMSeg generates predictions for 3D semantic segmentation in both the source and target domains, denoted as $\boldsymbol{\mathcal{P}}_{2 \mathrm{D}}^{\mathrm{S}}$, $\boldsymbol{\mathcal{P}}_{3\mathrm{D}}^{ \mathrm{S}}$, $\boldsymbol{\mathcal{P}}_{2 \mathrm{D}}^{\mathrm{T}}$, and $\boldsymbol{\mathcal{P}}_{3\mathrm{D}}^{ \mathrm{T}}$. 
Subsequently, the source domain predictions are supervised using the corresponding source domain labels, whereas our target domain predictions are supervised using accurate pseudo labels derived from our proposed VFM-PL method.
With the help of our proposed VFM-PL and FrustumMixing, the performance of cross-modal 3D UDA segmentation can be boosted.

\subsection{VFM-PL: Adapting Prior Knowledge of VFM}\label{subsec:leverageVFM}
SEEM has been trained on rich image-text pairs across numerous scenes. It has learned robust visual priors and can provide accurate object-level 2D masks with accurate semantic labels. In light of this, we introduce VFM-PL to generate refined pseudo labels. 
VFM-PL comprises two steps: 
(1) Pre-train a 2D neural network that could predict semantic labels on target domain; 
(2) Leverage SEEM~\cite{zou2023seem} and the pre-trained 2D neural network to generate pseudo labels for the subsequent training stage.

\noindent\textbf{Cross-modal and supervised pre-training.}
We follow the image and point cloud information flow in the framework depicted in Fig. \ref{fig:framework} (mixed data are excluded in this stage) to pre-train a 2D neural network. 
In both the 2D and 3D neural networks, there are two projection heads. The first projection head ($\mathbf{P_{H_1}}$) is specifically designed for the final prediction.
In the source domain, the labeled data can provide the first head of precise semantic labels for both neural networks and help them capture significant domain features. The target domain, on the other hand, provides no supervision for the first head in this stage. 
The second projection head ($\mathbf{P_{H_2}}$) is designed to transfer visual knowledge across two modalities via KL-Divergence. More specifically, the 2D to 3D and 3D to 2D information exchange can be described as following~\cite{jaritz2022cross}:
\begin{equation} \mathcal{L}_{2D\rightarrow3D} = D_{KL}(3D_{\mathbf{P_{H_1}}}\mid\mid2D_{\mathbf{P_{H_2}}}),
\end{equation}
\begin{equation} \mathcal{L}_{3D\rightarrow2D} = D_{KL}(2D_{\mathbf{P_{H_1}}}\mid\mid3D_{\mathbf{P_{H_2}}}),
\end{equation}
where $\mathcal{L}_{2D\rightarrow3D}$ and $\mathcal{L}_{3D\rightarrow2D}$ are cross-modal losses.

At the end of the pre-training stage, the final state, i.e. the last checkpoints, of the 2D and 3D neural networks are kept for the generation of pseudo labels in the target domain.

\noindent\textbf{VFM assisted refinement of pseudo labels.}
Pseudo labels generated by the pre-trained 2D neural networks are considered to be noisy and lack precision as shown in Fig.~\ref{teaser}a. 
Applying these inaccurate labels as supervision signals introduces intrinsic segmentation errors in our neural networks. In contrast, SEEM could produce consistent and relatively precise semantic masks with accurate labels. Hence, we propose to leverage the robust visual prior learned by SEEM to further refine the produced pseudo labels. The overall procedure is shown in Fig.~\ref{fig:seem_pl}. Firstly, we input target image to SEEM and it produces pixel-wise segmentation prediction. By applying softmax function to its output logits, we obtain the class-wise probability distribution for each pixel. Then we exploit this robust visual prior to refining generated pseudo labels by averaging the predicted probabilities from SEEM and pre-trained neural 2D network: 
\begin{equation}\small
\textbf{PL}_{\textbf{R}} = \textit{Max}(\textit{Softmax}(\boldsymbol{\overline{\mathcal{P}}}_{2D_{\text{pretrain}}}) + \textit{Softmax}(\boldsymbol{\overline{\mathcal{P}}}_{\text{SEEM}}))\end{equation}
where $\textbf{PL}_{\textbf{R}}$ represents the pixel-wise refined pseudo label, $\textit{Softmax}$ stands for the Softmax function. $\boldsymbol{\overline{\mathcal{P}}}_{2D_{\text{pretrain}}}$ is the predicted probability of pre-trained 2D neural network, while $\boldsymbol{\overline{\mathcal{P}}}_{\text{SEEM}}$ denotes the predicted probability of SEEM.

Although the outputs of pre-trained 2D neural network are imprecise, the domain-specific features are also captured during pre-training process. We argue that the learned yet noisy feature could help SEEM adapt its visual knowledge to the specific target domain which we are addressing. The empirical evaluation validates our assumption and we will elaborate on that later in Sec~\ref{sec:ablation}.

\begin{figure}[t]
	\centering
	\includegraphics[width=1\linewidth]{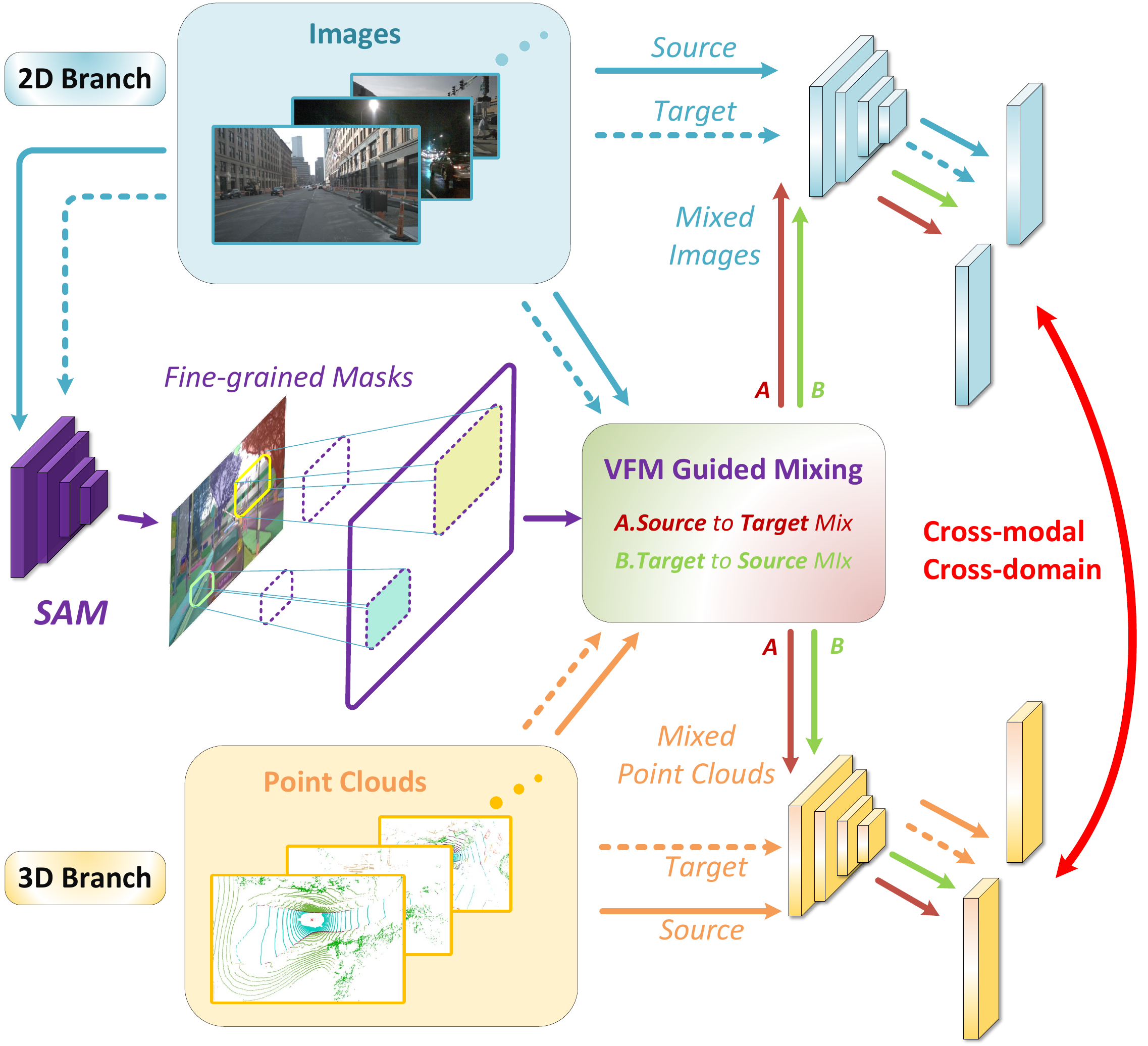}
	
	\caption{\textbf{FrustumMixing: VFM guided semantically mixing.} To further enhance the capability of neural networks to bridge the gap across domains, we propose to utilize SAM~\cite{Alexander2023sam} to generate fine-grained 2D masks by feeding images from both domains. The image mixing is realized by using masks that are generated according to one image to cut out corresponding areas, then fill in these masked areas with respective pixels selected from the other image.} 
	\label{fig:sam_mixing}
\end{figure}

\subsection{FrustumMixing: VFM guided Data Mixing}\label{subsec:frustrumMix}
To further facilitate the information exchange between different domains, we propose a new mixing strategy, FrustrumMixing. SAM~\cite{Alexander2023sam} has demonstrated its preeminent capability to generate precise yet fine-grained masks for various input images. These remarkable segmentation results have not only inspired us to develop this data-mixing approach but also provided us with the basic ingredients for mixing image and point cloud samples in a fine-grained manner. The overall FrustrumMixing pipeline is depicted in Fig.~\ref{fig:sam_mixing}. There are two mixing branches in our strategy, namely, the source-to-target mix and the target-to-source mix. The operation of both mixing branches is identical and the only difference is in the first step that which the image domain is selected to generate masks. Fig.~\ref{teaser}b demonstrates the target to source FrustrumMixing and we will explain this process as follows. Basically, there are four steps to perform the mixing. (1) Input target image to SAM and save the fine-grained masks. (2) Randomly sample a proportion of generated masks and merge them into one layer of mask. (3) Apply the fused mask to target image and paste the masked pixels onto source image and cover the original area. Now we have obtained the mixed \textbf{target to source} image sample. (4) The final step is to pick point clouds to construct mixed 3D data. By applying the 3D to 2D projection matrix, we could produce a point image that contains all necessary points within the sight of the 2D camera. The aligned 2D image and calculated point image now pave the way for applying the merged SAM mask to select point clouds. We utilize the same mask to choose points from the target point image and delete the points inside the corresponding area in the source point image. The mixed \textbf{target to source} point cloud sample is generated by filling up the emptied area in the source point image with picked points from the target domain. FrustrumMixing provides neural networks with semantically mixed samples from both domains and is beneficial for UDA 3D segmentation performance. We will analyze the effectiveness of this method in Sec.~\ref{sec:ablation}.

\section{Experiments}

\begin{table*}[ht]
	\centering
	\begin{tabular}{@{}lcccccccccccc}
		\toprule[1.5pt]
		\multirow{2}{*}{} & \multicolumn{3}{c}{A2D2/Sem.KITTI} 
		& \multicolumn{3}{c}{V.KITTI/S.KITTI}
		& \multicolumn{3}{c}{nuSc.L.Seg:USA/Sing.}
		& \multicolumn{3}{c}{nuSc.L.Seg:Day/Night}  \\
		\cmidrule(l{5pt}r{5pt}){2-4} \cmidrule(l{5pt}r{5pt}){5-7} \cmidrule(l{5pt}r{5pt}){8-10} \cmidrule(l{5pt}r{5pt}){11-13} 
		\multicolumn{1}{l}{Method} & 2D & 3D & Avg & 2D & 3D & Avg & 2D & 3D & Avg & 2D & 3D & Avg\\
		\midrule[1pt]
		Baseline (Source Only) 		 & 34.2 & 35.9 & \cellcolor{oceanicblue!20}40.4 & 26.8 & 42.0 & \cellcolor{oceanicblue!20}42.2 & 58.4 & 62.8 & \cellcolor{oceanicblue!20}68.2 & 47.8 & 68.8 & \cellcolor{oceanicblue!20}63.3 \\
		\midrule[1pt]
		xMUDA 		 & 38.6 & 45.8 & \cellcolor{oceanicblue!20}45.2 & 38.1 & 43.8 & \cellcolor{oceanicblue!20}44.7 & 64.1 & 62.4 & \cellcolor{oceanicblue!20}68.7 & 55.5 & 69.2 & \cellcolor{oceanicblue!20}\textbf{67.4} \\
		xMUDA$_{PL}$ & 41.2 & 49.8 & \cellcolor{oceanicblue!20}47.5 & 38.7 & \underline{46.1} & \cellcolor{oceanicblue!20}45.0 & 65.6 & 63.8 & \cellcolor{oceanicblue!20}68.4 & \underline{57.6} & 69.6 & \cellcolor{oceanicblue!20}64.4\\
		AUDA 	     & 43.0 & 43.6 & \cellcolor{oceanicblue!20}46.8 & 35.8 & 37.8 & \cellcolor{oceanicblue!20}41.3 & 64.0 & 64.0 & \cellcolor{oceanicblue!20}69.2 & 55.6 & \underline{69.8} & \cellcolor{oceanicblue!20}64.8\\		
		AUDA$_{PL}$  & \textbf{46.8} & 48.1 & \cellcolor{oceanicblue!20}50.6 & 35.9 & 45.5 & \cellcolor{oceanicblue!20}\underline{45.9} & \underline{65.9} & \underline{65.3} & \cellcolor{oceanicblue!20}\underline{70.6} & 54.3 & 69.6 & \cellcolor{oceanicblue!20}61.1\\
		DsCML+CMAL 	     & \underline{46.3} & 50.7 & \cellcolor{oceanicblue!20}\underline{51.0} & 38.4 & 38.4 & \cellcolor{oceanicblue!20}45.5 & 65.6 & 56.2 & \cellcolor{oceanicblue!20}66.1 & 50.9 & 49.3 & \cellcolor{oceanicblue!20}53.2\\	
		DsCML+CMAL $_{PL}$ & \textbf{46.8} & \underline{51.8} & \cellcolor{oceanicblue!20}\textbf{52.4} & \underline{39.6} & 41.8 & \cellcolor{oceanicblue!20}42.2 & 65.6 & 57.5 & \cellcolor{oceanicblue!20}66.9 & 51.4 & 49.8 & \cellcolor{oceanicblue!20}53.8\\	
		Ours 	     & 45.0 & \textbf{52.3} & \cellcolor{oceanicblue!20}50.0 & \textbf{57.2} & \textbf{52.0} & \cellcolor{oceanicblue!20}\textbf{61.0} & \textbf{70.0} & \textbf{65.6} & \cellcolor{oceanicblue!20}\textbf{72.3} &\textbf{60.6} & \textbf{70.5} & \cellcolor{oceanicblue!20}\underline{66.5}\\\midrule [1pt]\midrule[1pt]
		Oracle       & 59.3 & 71.9 & \cellcolor{oceanicblue!20}73.6 & 66.3 & 78.4 & \cellcolor{oceanicblue!20}80.1 & 75.4 & 76.0 & \cellcolor{oceanicblue!20}79.6 & 61.5 & 69.8 & \cellcolor{oceanicblue!20}69.2\\
		\bottomrule[1.5pt]
	\end{tabular}
    \caption{\textbf{Comparison of Cross-Modal Unsupervised Domain Adaptation for 3D Semantic Segmentation.} We report the mIoU results (with \textbf{best} and 2nd \underline{best})
on the target set for each network as well as the ensembling result by averaging the predicted probabilities from 2D and 3D network. Following experimental settings in~\cite{jaritz2022cross}, we compare methods (xMUDA~\cite{jaritz2022cross}, AUDA~\cite{liu2021adversarial}, DsCML~\cite{peng2021sparse}) that utilize 2D image and 3D points from both source and target domains. The `Baseline' model~\cite{jaritz2022cross} is trained on source domain $\boldsymbol{\mathcal{S}}$ only, which provides us the lower bound for UDA segmentation performance. The `Oracle'~\cite{jaritz2022cross} performs the assumed upper bound. It is not only trained on both domains, but also given the correct supervised label of target domain $\boldsymbol{\mathcal{T}}$.
Due to the lack of results in some settings from original papers of AUDA and DsCML, we utilize their published codes to produce corresponding results. For AUDA, only the results of A2D2/Sem.KITTI are available from the original paper. As to DsCML, we only utilize its code in V.KITTI/S.KITTI setting since the original paper reports the results of the other three settings. 
In most of the test scenarios, Our proposed method boosts the performance on segmentation task and achieves superior results when compared to other effective methods. Detailed analysis is provided in Sec.~\ref{sec:resAnalysis}. }
	\label{tab:mainResults}
\end{table*}

\subsection{Datasets}
To construct our domain adaptation scenarios, we utilized publicly available datasets including nuScenes-Lidarseg \cite{caesar2020nuscenes}, VirtualKITTI \cite{gaidon2016virtual}, SemanticKITTI \cite{behley2019semantickitti}, and A2D2 \cite{Geyer2020a2d2}. The details regarding the dataset splits can be found in the Appendix.
Our selected scenarios encompass various typical challenges in domain adaptation. These challenges include changes in scene layout, such as the transition between right-hand-side and left-hand-side driving in the nuScenes-Lidarseg: USA/Singapore scenario (\textbf{nuSc.L.Seg:USA/Sing.}). Additionally, we address lighting variations, such as the shift from day to night in the nuScenes-Lidarseg: Day/Night scenario (\textbf{nuSc.L.Seg:Day/Night}). Furthermore, we tackle the synthetic-to-real data shift by incorporating data from VirtualKITTI /SemanticKITTI (\textbf{V.KITTI/S.KITTI}), where we bridge the gap between simulated depth and RGB data to real LiDAR and camera data. Lastly, we explore different sensor setups and characteristics, such as resolution and FoV, through the A2D2/SemanticKITTI scenario (\textbf{A2D2/Sem.KITTI}).
Our code (\href{https://github.com/EtronTech/VFMSeg}{https://github.com/EtronTech/VFMSeg}) facilitates the replication of all training data and splits, and further details can be found in the Appendix.

\begin{figure*}[ht]
	\centering
	\includegraphics[width=1\linewidth]{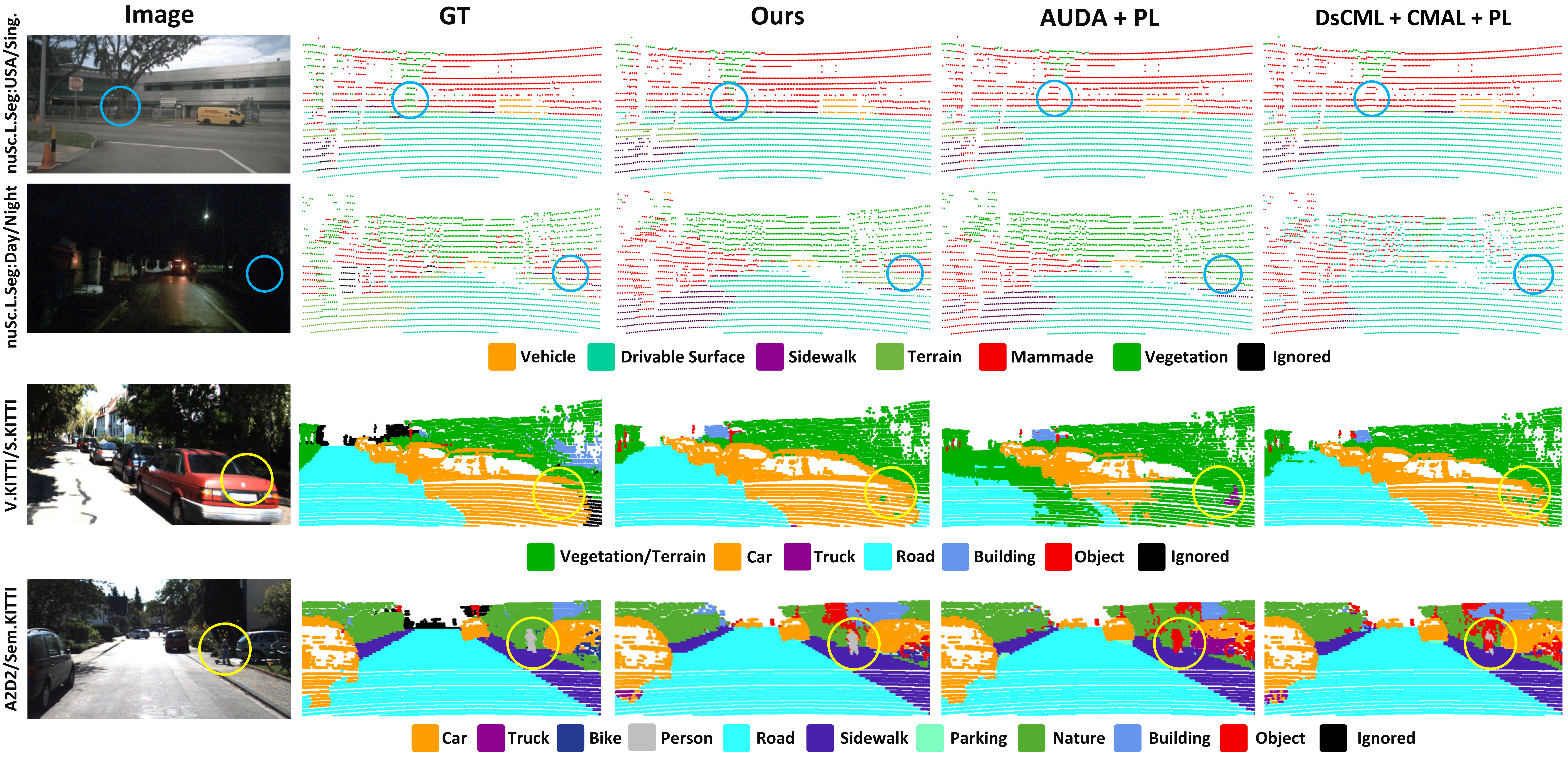}
	\caption{\textbf{Qualitative results.} We show the ensembling results of four scenarios by averaging the softmax outputs of 2D and 3D networks. Our method can improve the performance of 3D semantic segmentation. Noted that, by the merits of VFMs, our method can segment detailed objects very well. From top to bottom, the focused areas are the \underline{trunk} of a tree, \underline{manmade} objects under restricted lighting condition, the silhouette of a \underline{vehicle}, and most importantly, a \underline{kid} playing close to the road. }
	\label{fig:qa}
\end{figure*}

\subsection{Implementation Details}
\label{sec:implement_detail}

\noindent\textbf{Data Pre-processing.} Considering the computation resources required for VFMs and the repetitive nature of sampling data for training neural networks, we generate all masks for FrustrumMixing beforehand. Compared to generating semantic masks and fine-grained label-free masks on-the-fly, the training time in our hardware environment shrinks from weeks to days. For SEEM masks, we iterate all training samples and save both class labels and masked areas in a pickle file (.pkl). The process for SAM masks needs additional steps. The SAM mask data has the shape of an image and with only one channel to store `True' and `False', which indicates whether the respective pixel is masked. Since we perform a random sampling of SAM mask, as illustrated in Sec.~\ref{subsec:frustrumMix}, all fine-grained masks must be preserved for the training stage. However, the storage space required for all training images is enormous (estimated to be in a few Tera Bytes). Hence, we first give each mask a unique number and then merge all mask data into one matrix. Such a matrix is identical in size to the individual mask but stores the number instead. This pre-processing method simultaneously reduces the storage space and training time cost.

\noindent\textbf{Network Architecture.}
To ensure a fair comparison with the only existing multi-modal 3D domain adaptation method, we employ the following approaches:
For the 2D network, we utilize ResNet34~\cite{he2016deep}, which has been pre-trained on the ImageNet dataset, as the encoder for the U-Net~\cite{ronneberger2015u}.
For the 3D network, we employ SparseConvNet~\cite{graham20183d} with a U-Net architecture, implementing six rounds of down-sampling. Additionally, we adopt a voxel size of 5cm in the 3D network. This voxel size ensures that each voxel contains only one 3D point, maintaining a level of granularity suitable for the task.

\noindent\textbf{Training Details.}
Our method and the other baselines were trained and evaluated using the PyTorch toolbox on the Python 3.7 platform. The implementation of all proposed models was conducted on four NVIDIA RTX 3090Ti GPUs, each with 24GB of RAM.
During the training phase, we adopted a batch size of 8 and employed the Adam optimizer~\cite{kingma2014adam} with $\beta_1$ = 0.9 and $\beta_2$ = 0.999. The initial learning rate was set to 1e-3, and we utilized the poly learning rate policy~\cite{chen2017rethinking} with a power of 0.9. The maximum number of training iterations was set to 30k for \textbf{V.KITTI/S.KITTI}, the other three scenarios are ste to 100k.

\noindent\textbf{Evaluation.} Consistent with previous domain adaptation studies~\cite{jaritz2022cross, peng2021sparse}, we assess the performance of our model on the test set using the widely used PASCAL VOC intersection-over-union (IoU) metric. The mean IoU (mIoU) is calculated as the average of the IoU values across all categories.

\begin{table}
	\centering
	\begin{tabular}{@{}ccccc@{}}\toprule[1.5pt]
		\multirow{2}{*}{} & \multicolumn{4}{c}{Scenarios} \\
		\cmidrule(l{5pt}r{5pt}){2-5}
		\multicolumn{1}{c}{Method} & \#1 & \#2 & \#3 & \#4 \\
		\midrule[1pt] 
		Baseline (xMUDA)           &   38.6   &  38.1  &  64.1    &   55.5   \\\midrule[1pt]
		\multirow{2}{*}{}xMUDA$_{PL}$      &   \multirow{2}{*}{}  41.2  &  38.7  &   65.6   &   \underline{57.6}  \\
		\multicolumn{1}{c}{$\Delta$}       & 	$\uparrow$ 2.6  	  & $\uparrow$ 0.6  &   $\uparrow$ 1.5    &  $\uparrow$ 2.1   \\\midrule[1pt]
		\multirow{2}{*}{}SEEM Only         &   \multirow{2}{*}{}  35.7  &  51.3  &   50.5   &    33.7  \\
		\multicolumn{1}{c}{$\Delta$}	   & 	$\downarrow$ 2.9  	  & $\uparrow$ 13.2  &   $\downarrow$ 13.6   &   $\downarrow$ 21.8   \\\midrule[1pt]
		\multirow{2}{*}{}SEEM+2D Avg.      &   \multirow{2}{*}{}  \underline{43.0}  &  \underline{55.3}  &  \underline{ 67.7}   &   \textbf{57.9}  \\
		\multicolumn{1}{c}{$\Delta$}	   & 	 $\uparrow$ 4.4 	  &  $\uparrow$ 17.2 &   $\uparrow$ 3.6  &   $\uparrow$ 2.4  \\\midrule[1pt]
		\multirow{2}{*}{}VFM-PL      &   \multirow{2}{*}{} \textbf{43.6}   &  \textbf{55.7}  &   \textbf{68.8}   &   57.1  \\
		\multicolumn{1}{c}{$\Delta$}	   & 	 $\uparrow$ 5.0	  & $\uparrow$ 17.6  &   $\uparrow$ 4.7   &  $\uparrow$  1.6   \\
		\bottomrule[1.5pt]
	\end{tabular}
	\caption{\textbf{Ablation study} on the effect of pseudo-labels generated via VFM guidance. We report the mIoU segmentation performance of 2D networks to validate the effectiveness of proposed VFM-PL. Column \#1 to Column \#4 represents the \textbf{A2D2/Sem.KITTI}, \textbf{V.KITTI/S.KITTI}, \textbf{nuSc.L.Seg: USA/Singapore} and \textbf{nuSc.L.Seg: Day/Night} scenarios respectively. }
	\label{tab:ablationVFM}
\end{table}

\subsection{Experimental Results and Comparison}\label{sec:resAnalysis}
To validate the effectiveness of our proposed VFMSeg, we carried out four domain shift scenarios as introduced by~\cite{jaritz2022cross}.
Table~\ref{tab:mainResults} presents the experimental results and performance comparison of our method with previous unsupervised domain adaptation methods for 3D segmentation, following the setup introduced in Sec.~\ref{sec:implement_detail}. 
Each experiment includes two common reference methods: a baseline model called Source only, trained solely on the source domain, and an upper-bound model named Oracle, trained exclusively on the target data with annotations. 
And we compare our VFMSeg with other multi-modal methods based on xMUDA, such as AUDA~\cite{liu2021adversarial} and DsCML~\cite{peng2021sparse}.
Among these methods, xMUDA achieves better performance on \textbf{V.KITTI $\rightarrow$ S.KITTI} and \textbf{Day $\rightarrow$ Night}, while AUDA obtains comparable results on \textbf{USA $\rightarrow$ Singpore}.
By the merits of our VFM-PL and FrustumMixing, our VFMSeg outperforms these methods by $+32.9 \%$ (\textbf{V.KITTI $\rightarrow$ S.KITTI}), $+2.4 \%$ (\textbf{USA $\rightarrow$ Singpore}). For \textbf{Day $\rightarrow$ Night} scenario, VFMSeg achieves the best 3D segmentation performance and is even $0.7\%$ higher than the assumed upper bound, `Oracle' model, which is fully supervised on target domain $\boldsymbol{\mathcal{T}}$. As to \textbf{A2D2 $\rightarrow$ S.KITTI} scenario, that SEEM provides no class label near the semantic meaning of `Trunk' under this setting. Hence, we fully ignored this supervised signal for training and the noise introduced via this processing method could lead to the inferior results in 2D segmentation. Still, VFMSeg achieves the second best performance among all segmentation results in this setting and is only $0.1\%$ behind the best results ( $52.4\%$ from $\text{DsCML}+\text{CMAL}_{\text{PL}}$ ). Overall, the empirical experiments have validated the the effectiveness of our proposed VFMSeg method.

\begin{table}
	\centering
	\begin{tabular}{@{}c|c|ccc@{}}\toprule[1.5pt]

		\multirow{1}{*}{No.} & w. Mix &  2D & 3D & Avg \\ \midrule[1pt]

		\multirow{2}{*}{\#1}    &   \multirow{2}{*}{} \ding{56}  &  43.6  &   50.3   &   47.6 \\
							    & 	  \ding{52}	  &   45.0 (\textbf{+1.4})    &   52.3 (\textbf{+2.0})  & 50.0 (\textbf{+2.4}) \\\midrule[1pt]
		\multirow{2}{*}{\#2}    &   \multirow{2}{*}{} \ding{56}  &   55.7   &    49.9 & 59.8 \\
								& 	 \ding{52}	  &    57.2 (\textbf{+1.5})   &   51.9 (\textbf{+2.0})  & 61.0 (\textbf{+1.2}) \\\midrule[1pt]
		\multirow{2}{*}{\#3}    &   \multirow{2}{*}{}  \ding{56} &   68.2   &   64.0 & 71.1 \\
								& 	 \ding{52}  &   70.0 (\textbf{+1.8})  &    65.6 (\textbf{+1.6}) & 72.3 (\textbf{+1.2}) \\\midrule[1pt]
		\multirow{2}{*}{\#4}    &   \multirow{2}{*}{} \ding{56}   &   57.1  &   69.8 & 68.3\\
					   			& 	 \ding{52} &     60.6 (\textbf{+3.5})   &   70.5 (\textbf{+0.7})  & 66.5 (-1.8)\\
		\bottomrule[1.5pt]
	\end{tabular}
	\caption{\textbf{Ablation study} on the effect of mixing strategy under VFM guidance. Row \#1 to Row \#4 represents the \textbf{A2D2/Sem.KITTI}, \textbf{V.KITTI/S.KITTI}, \textbf{nuSc.L.Seg: USA/Singapore} and \textbf{nuSc.L.Seg: Day/Night} scenarios respectively. `w.Mix' indicates whether the mixed data is involved in the training process.}
	\label{tab:ablationMix}
\end{table}

\subsection{Ablation Study}
\label{sec:ablation}
To demonstrate the effectiveness of each module in our method, we conduct ablation studies on four unsupervised domain adaptation scenarios. Furthermore, we evaluate the impact of various visual foundation models on the performance of our method.

\begin{figure}
    \centering
    \includegraphics[width=\linewidth]{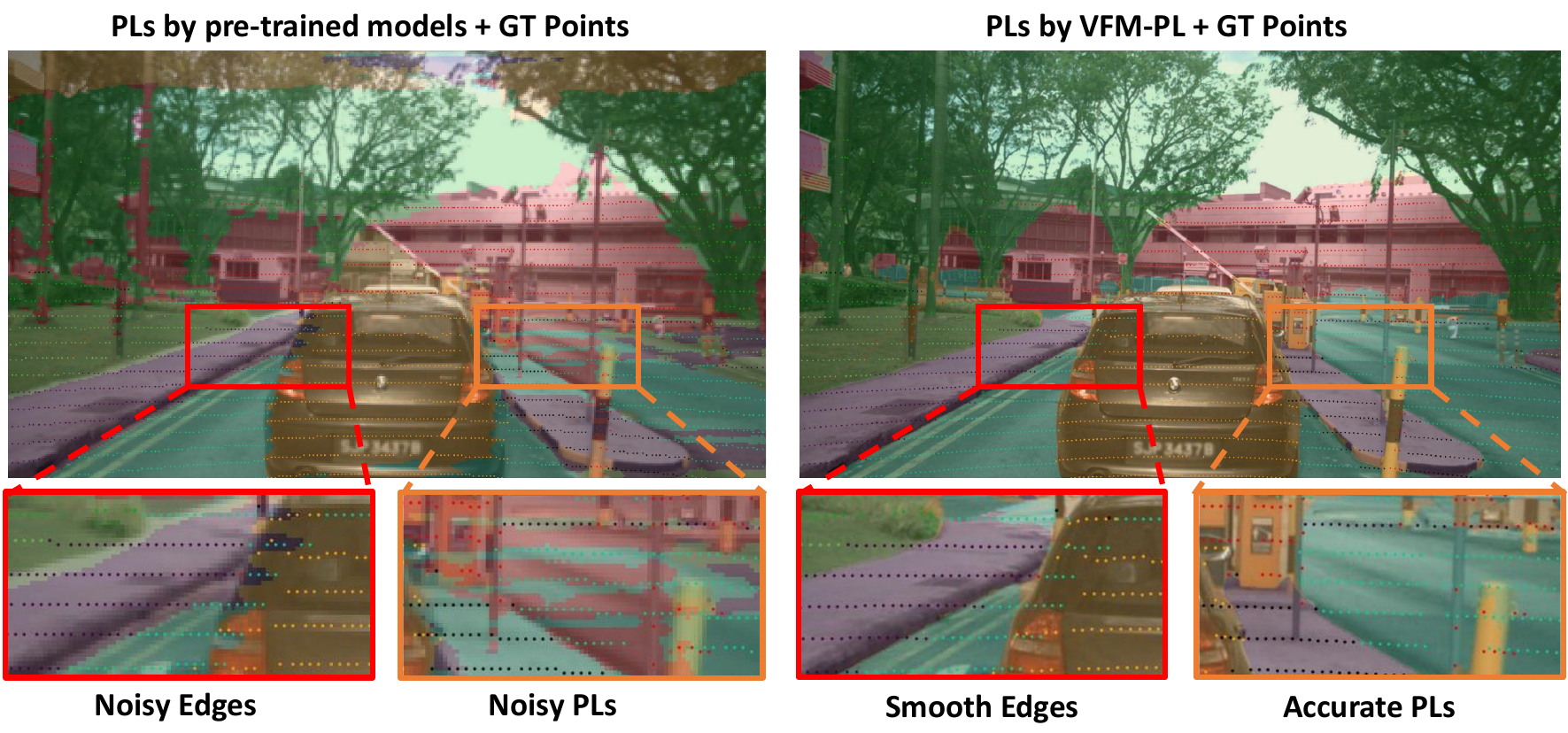}
    \caption{Projection errors caused by projecting points onto images. PLs from the pre-trained model tend to be noisy but can learn the noisy edges from projection errors. Our VFM-PL is able to generate accurate PLs, where the smooth edges will cause gaps compared with ground truth.}
    \label{fig:ablation1}
\end{figure}

\noindent\textbf{Effects of VFM-guided Accurate Pseudo-Label Generation.}
To validate the effectiveness of our proposed VFM-PL, further ablation studies are conducted.
Table~\ref{tab:ablationVFM} demonstrates that fine-tuning the xMUDA model using pseudo labels generated by pre-trained xMUDA models results in a marginal improvement in performance across all four UDA scenarios.
Surprisingly, our findings indicate that the pseudo labels generated by SEEM only outperform xMUDA$_{PL}$ in the VirtualKITTI/SemanticKITTI setting.
This observation can be primarily attributed to the presence of projection errors in the point clouds when projected onto images.
From Fig.~\ref{fig:ablation1}, it is evident that the edges of objects in the ground truth exhibit noise, whereas the pseudo-labels generated by our VFM-PL demonstrate remarkably smooth edges.
Therefore, we employ the pseudo-labels generated by pre-trained models to assist VFM-PL in bridging the gap between the pseudo-labels obtained from SEEM and the ground truth.
To achieve this, we perform an ensemble of the pseudo-labels obtained from pre-trained models and those from SEEM by averaging the softmax logits. This approach enables the supervision of the noisy edges in the ground truth through the pseudo-labels from pre-trained models, while the main parts of objects can be learned from the pseudo-labels generated by SEEM.

\noindent\textbf{Effects of VFM-guided Semantic Data Augmentation}
To gain a deeper understanding of the effectiveness of our FrustumMixing method, we conducted additional ablation studies. 
Table~\ref{tab:ablationMix} demonstrates that models trained with our VFM-guided semantic data augmentation exhibit significant improvements in both 2D and 3D performance across all four UDA scenarios, thereby leading to better improvements in average performance.
The results obtained from our experiments clearly indicate that our FrustumMixing approach, guided by masks generated by VFM, operates similarly to the concept of view frustum. This results in more effective semantic data augmentation, as opposed to a random mix-up of source and target samples.
The incorporation of semantic data augmentation contributes to improving the learning process of the networks, ultimately leading to enhanced overall performance.

\section{Discussion and Future Work}
The robust and consistent visual priors of VFMs inspired us to leverage their capability to facilitate our 3D segmentation task. To the best of our knowledge, we are the \textbf{first} to incorporate \textbf{two} VFMs into UDA for 3D framework. The key takeaway here is plain and simple. Feeding neural networks with semantically mixed samples across various domains is foreseeingly beneficial. The fine-grained, yet rich in visual semantic meaning, masks generated by SAM fit right on the spot for generating sufficiently mixed samples. Besides, the lack of object-level text labels in SAM masks could be compensated by adopting segmentation VFMs that are trained with abundant image-text pairs, in our case, we choose to utilize SEEM for refining pseudo labels for the target domain.

Projection errors are a common issue encountered in different cross-modal autonomous driving datasets, creating a challenge for cross-modal UDA in 3D semantic segmentation. Although our VPM-PL approach, as discussed in Sec.~\ref{sec:ablation}, helps alleviate this problem, it does not completely solve it. As a result, our future work will concentrate on addressing projection errors that arise when projecting point clouds onto images. Once this issue is effectively resolved, it has the potential to further enhance the performance of 3D semantic segmentation.

\section{Conclusion}

This paper introduces VFMSeg as a novel framework that boosts cross-modal unsupervised domain adaptation for 3D semantic segmentation by the merits of VFMs.
VFMSeg addresses two primary challenges encountered by prior research.
Firstly, to tackle the noisy pseudo labels generated from models trained on labeled source domain and unlabeled target domain, we exploit the knowledge prior from VFM to obtain more accurate pseudo labels for the target domain.
Secondly, to narrow the domain gap, our FrustumMixing leverages another VFM to mix the point clouds and images from both domains. Such fine-grained source-target mixing can be regarded as a potent augmentation technique, which effectively minimizes the domain gap.
Extensive experiments were conducted under several UDA scenarios, demonstrating that our VFMSeg outperforms all the compared state-of-the-art methods by a significant margin.

{
    \small
    \bibliographystyle{ieeenat_fullname}
    \bibliography{ref}
}

\clearpage
\setcounter{page}{1}
\maketitlesupplementary
\vspace{0.5cm}

In this appendix, we supplement more content from the following aspects to support the proposed VFMSeg in the main body of this paper:

\begin{itemize}
	\item Sec.~\ref{apnd:pseudocode} elaborates on the proposed VFMSeg via pseudo codes.
	\item Sec.~\ref{apnd:dataset} provides more details about experimental settings and the splits of each dataset.
	\item Sec.~\ref{apnd:additionalExp} gives additional comparison analysis of VFMSeg, including an alternative method for FrustrumMixing.
        \item Sec.~\ref{apnd:vfmcomparison} demonstrates the difference between SAM and SEEM generated masks.
	\item Sec.~\ref{apnd:viz} gives more visualization results for four experiment settings.
\end{itemize}

\section{Pseudo Code for the Proposed VFMSeg}
\label{apnd:pseudocode}
There are two learning stages in VFMSeg. The first stage is for pre-training a 2D and 3D neural network that could provide pseudo labels with limited precision (\underline{Algorithm ~\ref{Algo-VFM-PL}} gives pseudo code of VFM-PL). For the target 3D domain, we apply a pre-trained 3D network to predict labels. For the target 2D domain, SEEM is utilized in conjunction with the pre-trained 2D network to provide more accurate pseudo labels.

As to the second learning stage, we first perform FrustrumMixing via sampled SAM masks to generate fine-grained and rich in semantic mixed samples to further bridge the gap between two domains (\underline{Algorithm~\ref{Algo-FrustrumMixing}} provides pseudo code for detailed mixing procedure). After we obtained the mixed source-to-target and target-to-source samples, a 2D and 3D neural network leverages these training samples to optimize their parameters for 3D segmentation task (elaborates in \underline{Algorithm~\ref{Algo-Training}}).

\begin{algorithm}[ht]
\KwIn{\underline{Source} images, point clouds, labels and indices: \\
$\boldsymbol{\mathcal{D}_{S}} = \{ \boldsymbol{\mathcal{S}}_{\text{2D}}^{(i)}, \boldsymbol{\mathcal{S}}_{\text{3D}}^{(i)}, \boldsymbol{\mathcal{S}}_{\text{Labels}}^{(i)}, \boldsymbol{\mathcal{S}}_{\text{Indices}}^{(i)}, i \in (1,2,...,I)\}$. \underline{Target} images, point clouds and indices: $\boldsymbol{\mathcal{D}_{T}} = \{ \boldsymbol{\mathcal{T}}_{\text{2D}}^{(j)}, \boldsymbol{\mathcal{T}}_{\text{3D}}^{(j)}, \boldsymbol{\mathcal{T}}_{\text{Indices}}^{(i)}, j \in (1,2,...,J)\}$. Maximum iteration $\boldsymbol{\mathcal{N}}$. 2D and 3D neural network $\boldsymbol{\mathcal{G}_{\text{2D}}}$, $\boldsymbol{\mathcal{G}_{\text{3D}}}$ (both neural networks have two projection heads, $\mathbf{P_{H_1}}$ and $\mathbf{P_{H_2}}$, their predictions are annotated as $\boldsymbol{\overline{\mathcal{P}}}$ and $\boldsymbol{\overline{\mathcal{P'}}}$ respectively). The SEEM model $\boldsymbol{\mathcal{G}_{SEEM}}$. Cross-modal weights for source and target domain, $\lambda_{\text{xm.src}}$, $\lambda_{\text{xm.trg}}$. Batch size $B$.}
	
	\Repeat{Reach the maximum iteration $\boldsymbol{\mathcal{N}}$}
	{
		Sample one batch of data from $\boldsymbol{\mathcal{D}_{S}}$ and $\boldsymbol{\mathcal{D}_{T}}$ \\
		$\boldsymbol{\overline{\mathcal{P}}}_{\text{2D}_{\text{src}}}, \boldsymbol{\overline{\mathcal{P'}}}_{\text{2D}_{\text{src}}} = \text{Sample}(\boldsymbol{\mathcal{G}_{\text{2D}}}(\boldsymbol{\mathcal{S}}_{\text{2D}}^{B}), \boldsymbol{\mathcal{S}}_{\text{Indices}}^{B})$

		$\boldsymbol{\overline{\mathcal{P}}}_{\text{3D}_{\text{src}}}, \boldsymbol{\overline{\mathcal{P'}}}_{\text{3D}_{\text{src}}} = \text{Sample}(\boldsymbol{\mathcal{G}_{\text{3D}}}(\boldsymbol{\mathcal{S}}_{\text{3D}}^{B}), \boldsymbol{\mathcal{S}}_{\text{Indices}}^{B})$

		$\boldsymbol{\overline{\mathcal{P}}}_{\text{2D}_{\text{trg}}}, \boldsymbol{\overline{\mathcal{P'}}}_{\text{2D}_{\text{trg}}} = \text{Sample}(\boldsymbol{\mathcal{G}_{\text{2D}}}(\boldsymbol{\mathcal{T}}_{\text{2D}}^{B}), \boldsymbol{\mathcal{T}}_{\text{Indices}}^{B})$

		$\boldsymbol{\overline{\mathcal{P}}}_{\text{3D}_{\text{trg}}}, \boldsymbol{\overline{\mathcal{P'}}}_{\text{3D}_{\text{trg}}} = \text{Sample}(\boldsymbol{\mathcal{G}_{\text{3D}}}(\boldsymbol{\mathcal{T}}_{\text{3D}}^{B}), \boldsymbol{\mathcal{T}}_{\text{Indices}}^{B})$
		
		$L^{\text{src}}_{\text{2D}} = \text{CrossEntropy}(\boldsymbol{\overline{\mathcal{P}}}_{\text{2D}_{\text{src}}}, \boldsymbol{\mathcal{S}}_{\text{Labels}}^{B})$
		
		$L^{\text{src}}_{\text{3D}} = \text{CrossEntropy}(\boldsymbol{\overline{\mathcal{P}}}_{\text{3D}_{\text{src}}}, \boldsymbol{\mathcal{S}}_{\text{Labels}}^{B})$
		
		$L^{\text{src}}_{\text{2D}\rightarrow\text{3D}} = \text{KL-Divergence}(\boldsymbol{\overline{\mathcal{P'}}}_{\text{3D}_{\text{src}}}, \boldsymbol{\overline{\mathcal{P}}}_{\text{2D}_{\text{src}}})$

		$L^{\text{src}}_{\text{3D}\rightarrow\text{2D}} = \text{KL-Divergence}(\boldsymbol{\overline{\mathcal{P'}}}_{\text{2D}_{\text{src}}}, \boldsymbol{\overline{\mathcal{P}}}_{\text{3D}_{\text{src}}})$

		$L^{\text{trg}}_{\text{2D}\rightarrow\text{3D}} = \text{KL-Divergence}(\boldsymbol{\overline{\mathcal{P'}}}_{\text{3D}_{\text{trg}}}, \boldsymbol{\overline{\mathcal{P}}}_{\text{2D}_{\text{trg}}})$

		$L^{\text{trg}}_{\text{3D}\rightarrow\text{2D}} = \text{KL-Divergence}(\boldsymbol{\overline{\mathcal{P'}}}_{\text{2D}_{\text{trg}}}, \boldsymbol{\overline{\mathcal{P}}}_{\text{3D}_{\text{trg}}})$
		
		Backward($L^{\text{src}}_{\text{2D}}$), Backward($L^{\text{src}}_{\text{3D}}$)\\
		Backward($\lambda_{\text{xm.src}}L^{\text{src}}_{\text{2D}\rightarrow\text{3D}}$), Backward($\lambda_{\text{xm.src}}L^{\text{src}}_{\text{3D}\rightarrow\text{2D}}$)\\
		Backward($\lambda_{\text{xm.trg}}L^{\text{trg}}_{\text{2D}\rightarrow\text{3D}}$), Backward($\lambda_{\text{xm.trg}}L^{\text{trg}}_{\text{3D}\rightarrow\text{2D}}$)\\
		Update($\boldsymbol{\mathcal{G}_{\text{2D}}}$), Update($\boldsymbol{\mathcal{G}_{\text{3D}}}$)
		
	} 
	
	\ForEach{$j \in \{1,2,...,J \}$}{
	$\boldsymbol{\overline{\mathcal{P}}}_{\text{2D}} = \boldsymbol{\mathcal{G}_{\text{2D}.\mathbf{P_{H_1}}}}(\boldsymbol{\mathcal{T}}_{\text{2D}}^{(j)})$

	$\boldsymbol{\overline{\mathcal{P}}}_{\text{SEEM}} = \boldsymbol{\mathcal{G}_{SEEM}} (\boldsymbol{\mathcal{T}}_{\text{2D}}^{(j)})$

	$\boldsymbol{\overline{\mathcal{P}}}_{\text{3D}} = \boldsymbol{\mathcal{G}_{\text{3D}.\mathbf{P_{H_1}}}}(\boldsymbol{\mathcal{T}}_{\text{3D}}^{(j)})$
	
	$\textbf{PL}^{(j)}_{\textbf{R}} = \textit{Max}(\textit{Softmax}(\boldsymbol{\overline{\mathcal{P}}}_{\text{2D}}) + \textit{Softmax}(\boldsymbol{\overline{\mathcal{P}}}_{\text{SEEM}}))$
	
	$\textbf{PL}^{(j)}_{\textbf{3D}} = \textit{Max}(\textit{Softmax}(\boldsymbol{\overline{\mathcal{P}}}_{\text{3D}}))$
	
	}
\KwOut{Pseudo labels for target domain: \\
    $\boldsymbol{\mathcal{T}}_{\text{PL}} = \{\textbf{PL}^{(j)}_{\textbf{R}}, \textbf{PL}^{(j)}_{\textbf{3D}}, j = (1,2...,J) \}$}
	\caption{VFM guided pseudo label generation (VFM-PL). }
\label{Algo-VFM-PL}
\end{algorithm}

\begin{algorithm}[ht]\small
	\KwIn{\underline{Source} images, point clouds, corresponding \\labels and indices : $\boldsymbol{\mathcal{D}_{S}} = \{ \boldsymbol{\mathcal{S}}_{\text{2D}}^{(i)}, \boldsymbol{\mathcal{S}}_{\text{3D}}^{(i)}, \boldsymbol{\mathcal{S}}_{\text{Labels}}^{(i)}, \boldsymbol{\mathcal{S}}_{\text{Indices}}^{(i)}, i \in (1,2,...,I)\}$. \underline{Target} images, point clouds, SEEM generated pseudo labels (VFM-PL) and indices:  $\boldsymbol{\mathcal{D}_{T}} = \{ \boldsymbol{\mathcal{T}}_{\text{2D}}^{(j)}, \boldsymbol{\mathcal{T}}_{\text{3D}}^{(j)}, \boldsymbol{\mathcal{T}}_{\text{PL}}^{(j)},\boldsymbol{\mathcal{T}}_{\text{Indices}}^{(j)}, j \in (1,2,...,J)\}$. The SAM model $\boldsymbol{\mathcal{G}_{SAM}}$ and sample ratio $r$.
	}
	
	\ForEach{Input data sample $\boldsymbol{\mathcal{D}^{i}_{\boldsymbol{\mathcal{S}}}}$ and $\boldsymbol{\mathcal{D}^{j}_{\boldsymbol{\mathcal{T}}}}$}{
		$\text{Mask}_{\text{src}} = \text{RandomSample}(\boldsymbol{\mathcal{G}_{SAM}}(\boldsymbol{\mathcal{S}}_{\text{2D}}^{(i)}),r)$
		
		$\text{Mask}_{\text{trg}} = \text{RandomSample}(\boldsymbol{\mathcal{G}_{SAM}}(\boldsymbol{\mathcal{T}}_{\text{2D}}^{(j)}),r)$

		$\text{RemainderMask}_{\text{src}} = \text{FullSize} - \text{Mask}_{\text{src}}$
		
		$\text{RemainderMask}_{\text{trg}} = \text{FullSize} - \text{Mask}_{\text{trg}}$
		
		$\text{Image}^{1}_{\text{src}} = \text{Apply}(\text{Mask}_{\text{src}},\boldsymbol{\mathcal{S}}_{\text{2D}}^{(i)})$
		
		$\text{Image}^{1}_{\text{trg}} = \text{Apply}(\text{RemainderMask}_{\text{src}},\boldsymbol{\mathcal{T}}_{\text{2D}}^{(j)})$
		
		$\textbf{Mix}^{\text{2D}}_{\text{src}\rightarrow\text{trg}} = \text{Image}^{1}_{\text{src}} + \text{Image}^{1}_{\text{trg}}$
		
		$\text{Image}^{2}_{\text{trg}} = \text{Apply}(\text{Mask}_{\text{trg}},\boldsymbol{\mathcal{T}}_{\text{2D}}^{(j)})$
		
		$\text{Image}^{2}_{\text{src}} = \text{Apply}(\text{RemainderMask}_{\text{trg}},\boldsymbol{\mathcal{S}}_{\text{2D}}^{(i)})$
		
		$\textbf{Mix}^{\text{2D}}_{\text{trg}\rightarrow\text{src}} = \text{Image}^{2}_{\text{src}} + \text{Image}^{2}_{\text{trg}}$
		
		$\text{Labels}^{1}_{\text{src}} = \text{Apply}(\text{Mask}_{\text{src}},\boldsymbol{\mathcal{S}}_{\text{Labels}}^{(i)})$
		
		$\text{Labels}^{1}_{\text{trg}} = \text{Apply}(\text{RemainderMask}_{\text{src}},\boldsymbol{\mathcal{T}}_{\text{PL}}^{(j)})$
		
		$\textbf{Mix}^{\text{Labels}}_{\text{src}\rightarrow\text{trg}} = \text{Labels}^{1}_{\text{src}} + \text{Labels}^{1}_{\text{trg}}$
		
		$\text{Labels}^{2}_{\text{trg}} = \text{Apply}(\text{Mask}_{\text{trg}},\boldsymbol{\mathcal{T}}_{\text{PL}}^{(j)})$
		
		$\text{Labels}^{2}_{\text{src}} = \text{Apply}(\text{RemainderMask}_{\text{trg}},\boldsymbol{\mathcal{S}}_{\text{Labels}}^{(i)})$
		
		$\textbf{Mix}^{\text{Labels}}_{\text{trg}\rightarrow\text{src}} = \text{Labels}^{2}_{\text{src}} + \text{Labels}^{2}_{\text{trg}}$
		
		$\text{Indices}^{1}_{\text{src}} = \text{Apply}(\text{Mask}_{\text{src}},\boldsymbol{\mathcal{S}}_{\text{Indices}}^{(i)})$
		
		$\text{Indices}^{1}_{\text{trg}} = \text{Apply}(\text{RemainderMask}_{\text{src}},\boldsymbol{\mathcal{T}}_{\text{Indices}}^{(j)})$
		
		$\textbf{Mix}^{\text{Indices}}_{\text{src}\rightarrow\text{trg}} = \text{Indices}^{1}_{\text{src}} + \text{Indices}^{1}_{\text{trg}}$
		
		$\text{Indices}^{2}_{\text{trg}} = \text{Apply}(\text{Mask}_{\text{trg}},\boldsymbol{\mathcal{T}}_{\text{Indices}}^{(j)})$
		
		$\text{Indices}^{2}_{\text{src}} = \text{Apply}(\text{RemainderMask}_{\text{trg}},\boldsymbol{\mathcal{S}}_{\text{Indices}}^{(i)})$
		
		$\textbf{Mix}^{\text{Indices}}_{\text{trg}\rightarrow\text{src}} = \text{Indices}^{2}_{\text{src}} + \text{Indices}^{2}_{\text{trg}}$
		
		$\text{PointImage}^{1}_{\text{src}} = \text{Apply}(\text{Mask}_{\text{src}},\boldsymbol{\mathcal{S}}_{\text{3D}}^{(i)})$
		
		$\text{PointImage}^{1}_{\text{trg}} = \text{Apply}(\text{RemainderMask}_{\text{src}},\boldsymbol{\mathcal{T}}_{\text{3D}}^{(j)})$
		
		$\textbf{Mix}^{\text{3D}}_{\text{src}\rightarrow\text{trg}} = \text{PointImage}^{1}_{\text{src}} + \text{PointImage}^{1}_{\text{trg}}$
		
		$\text{PointImage}^{2}_{\text{trg}} = \text{Apply}(\text{Mask}_{\text{trg}},\boldsymbol{\mathcal{T}}_{\text{3D}}^{(j)})$
		
		$\text{PointImage}^{2}_{\text{src}} = \text{Apply}(\text{RemainderMask}_{\text{trg}},\boldsymbol{\mathcal{S}}_{\text{3D}}^{(i)})$
		
		$\textbf{Mix}^{\text{3D}}_{\text{trg}\rightarrow\text{src}} = \text{PointImage}^{2}_{\text{src}} + \text{PointImage}^{2}_{\text{trg}}$
		
	}
	
	\KwOut{Mixed data samples in domain $\boldsymbol{\mathcal{M}}$:
		
		$\boldsymbol{\mathcal{D}_{\boldsymbol{\mathcal{M}}}} = \{ \textbf{Mix}^{\text{2D}}_{\text{src}\rightarrow\text{trg}},  \textbf{Mix}^{\text{2D}}_{\text{trg}\rightarrow\text{src}}, \textbf{Mix}^{\text{3D}}_{\text{src}\rightarrow\text{trg}}, \textbf{Mix}^{\text{3D}}_{\text{trg}\rightarrow\text{src}},$\\
		$\textbf{Mix}^{\text{Labels}}_{\text{src}\rightarrow\text{trg}},\textbf{Mix}^{\text{Labels}}_{\text{trg}\rightarrow\text{src}},\textbf{Mix}^{\text{Indices}}_{\text{src}\rightarrow\text{trg}}, \textbf{Mix}^{\text{Indices}}_{\text{trg}\rightarrow\text{src}}  \}$
	}
	\caption{FrustrumMixing.}
\label{Algo-FrustrumMixing}
\end{algorithm}

\begin{algorithm}[ht]\footnotesize\label{pseudocode:training}
	\KwIn{\underline{Source} images, point clouds and supervised\\
	class labels: $\boldsymbol{\mathcal{D}_{S}} = \{ \boldsymbol{\mathcal{S}}_{\text{2D}}^{(i)}, \boldsymbol{\mathcal{S}}_{\text{3D}}^{(i)}, \boldsymbol{\mathcal{S}}_{\text{Labels}}^{(i)}, \boldsymbol{\mathcal{S}}_{\text{Indices}}^{(i)}, i \in (1,2,...,I)\}$. \underline{Target} images, point clouds and SEEM generated pseudo labels (VFM-PL): \\ $\boldsymbol{\mathcal{D}_{T}} = \{ \boldsymbol{\mathcal{T}}_{\text{2D}}^{(j)}, \boldsymbol{\mathcal{T}}_{\text{3D}}^{(j)}, \boldsymbol{\mathcal{T}}_{\text{PL}}^{(j)}, \boldsymbol{\mathcal{T}}_{\text{Indices}}^{(j)}, j \in (1,2,...,J)$\\$\}$. Maximum iteration $\boldsymbol{\mathcal{N}}$. 2D neural network $\boldsymbol{\mathcal{G}_{\text{2D}}}$, 3D neural network $\boldsymbol{\mathcal{G}_{\text{3D}}}$. Cross-modal weights for source, target and mixed domain, $\lambda_{\text{xm.src}}$, $\lambda_{\text{xm.trg}}$ and $\lambda_{\text{xm.m}}$. Batch size $B$.
	
	
	}
	
	\Repeat{Reach the maximum iteration $\boldsymbol{\mathcal{N}}$}
{
	Sample one batch of data from $\boldsymbol{\mathcal{D}_{S}}$ and $\boldsymbol{\mathcal{D}_{T}}$ \\
	\ForEach{Data sample whitin the batch}{
	$\boldsymbol{\mathcal{D}^{b}_{\boldsymbol{\mathcal{M}}}} = \text{FrustrumMixing}(\boldsymbol{\mathcal{D}^{b}_{\boldsymbol{\mathcal{S}}}},\boldsymbol{\mathcal{D}^{b}_{\boldsymbol{\mathcal{T}}}})$
	}
	
	$\boldsymbol{\overline{\mathcal{P}}}_{\text{2D}_{\text{src}}}, \boldsymbol{\overline{\mathcal{P'}}}_{\text{2D}_{\text{src}}} = \text{Sample}(\boldsymbol{\mathcal{G}_{\text{2D}}}(\boldsymbol{\mathcal{S}}_{\text{2D}}^{B}), \boldsymbol{\mathcal{S}}_{\text{Indices}}^{B})$

	$\boldsymbol{\overline{\mathcal{P}}}_{\text{3D}_{\text{src}}}, \boldsymbol{\overline{\mathcal{P'}}}_{\text{3D}_{\text{src}}} = \text{Sample}(\boldsymbol{\mathcal{G}_{\text{3D}}}(\boldsymbol{\mathcal{S}}_{\text{3D}}^{B}), \boldsymbol{\mathcal{S}}_{\text{Indices}}^{B})$

	$\boldsymbol{\overline{\mathcal{P}}}_{\text{2D}_{\text{trg}}}, \boldsymbol{\overline{\mathcal{P'}}}_{\text{2D}_{\text{trg}}} = \text{Sample}(\boldsymbol{\mathcal{G}_{\text{2D}}}(\boldsymbol{\mathcal{T}}_{\text{2D}}^{B}), \boldsymbol{\mathcal{T}}_{\text{Indices}}^{B})$

	$\boldsymbol{\overline{\mathcal{P}}}_{\text{3D}_{\text{trg}}}, \boldsymbol{\overline{\mathcal{P'}}}_{\text{3D}_{\text{trg}}} = \text{Sample}(\boldsymbol{\mathcal{G}_{\text{3D}}}(\boldsymbol{\mathcal{T}}_{\text{3D}}^{B}), \boldsymbol{\mathcal{T}}_{\text{Indices}}^{B})$

	$\boldsymbol{\overline{\mathcal{P}}}^{\text{Mix.1}}_{\text{2D}}, \boldsymbol{\overline{\mathcal{P'}}}^{\text{Mix.1}}_{\text{2D}} = \text{Sample}(\boldsymbol{\mathcal{G}_{\text{2D}}}(\boldsymbol{\mathcal{M}}_{\text{2D}_{\text{src}\rightarrow\text{trg}}}^{B}), \boldsymbol{\mathcal{M}}_{\text{Indices}_{\text{src}\rightarrow\text{trg}}}^{B})$

	$\boldsymbol{\overline{\mathcal{P}}}^{\text{Mix.1}}_{\text{3D}}, \boldsymbol{\overline{\mathcal{P'}}}^{\text{Mix.1}}_{\text{3D}} = \text{Sample}(\boldsymbol{\mathcal{G}_{\text{3D}}}(\boldsymbol{\mathcal{M}}_{\text{3D}_{\text{src}\rightarrow\text{trg}}}^{B}), \boldsymbol{\mathcal{M}}_{\text{Indices}_{\text{src}\rightarrow\text{trg}}}^{B})$

	$\boldsymbol{\overline{\mathcal{P}}}^{\text{Mix.2}}_{\text{2D}}, \boldsymbol{\overline{\mathcal{P'}}}^{\text{Mix.2}}_{\text{2D}} = \text{Sample}(\boldsymbol{\mathcal{G}_{\text{2D}}}(\boldsymbol{\mathcal{M}}_{\text{2D}_{\text{trg}\rightarrow\text{src}}}^{B}), \boldsymbol{\mathcal{M}}_{\text{Indices}_{\text{trg}\rightarrow\text{src}}}^{B})$

	$\boldsymbol{\overline{\mathcal{P}}}^{\text{Mix.2}}_{\text{3D}}, \boldsymbol{\overline{\mathcal{P'}}}^{\text{Mix.2}}_{\text{3D}} = \text{Sample}(\boldsymbol{\mathcal{G}_{\text{3D}}}(\boldsymbol{\mathcal{M}}_{\text{3D}_{\text{trg}\rightarrow\text{src}}}^{B}), \boldsymbol{\mathcal{M}}_{\text{Indices}_{\text{trg}\rightarrow\text{src}}}^{B})$

	$L^{\text{src}}_{\text{2D}} = \text{CrossEntropy}(\boldsymbol{\overline{\mathcal{P}}}_{\text{2D}_{\text{src}}}, \boldsymbol{\mathcal{S}}_{\text{Labels}}^{B})$

	$L^{\text{src}}_{\text{3D}} = \text{CrossEntropy}(\boldsymbol{\overline{\mathcal{P}}}_{\text{3D}_{\text{src}}}, \boldsymbol{\mathcal{S}}_{\text{Labels}}^{B})$

	$L^{\text{trg}}_{\text{2D}} = \text{CrossEntropy}(\boldsymbol{\overline{\mathcal{P}}}_{\text{2D}_{\text{trg}}}, \boldsymbol{\mathcal{T}}_{\text{PL}}^{B})$

	$L^{\text{trg}}_{\text{3D}} = \text{CrossEntropy}(\boldsymbol{\overline{\mathcal{P}}}_{\text{3D}_{\text{trg}}}, \boldsymbol{\mathcal{T}}_{\text{PL}}^{B})$

	$L^{\text{src}\rightarrow\text{trg}}_{\text{Mix.2D}} = \text{CrossEntropy}(\boldsymbol{\overline{\mathcal{P}}}^{\text{Mix.1}}_{\text{2D}}, \boldsymbol{\mathcal{M}}_{\text{PL}_{\text{src}\rightarrow\text{trg}}}^{B})$

	$L^{\text{src}\rightarrow\text{trg}}_{\text{Mix.3D}} = \text{CrossEntropy}(\boldsymbol{\overline{\mathcal{P}}}^{\text{Mix.1}}_{\text{3D}}, \boldsymbol{\mathcal{M}}_{\text{PL}_{\text{src}\rightarrow\text{trg}}}^{B})$

	$L^{\text{trg}\rightarrow\text{src}}_{\text{Mix.2D}} = \text{CrossEntropy}(\boldsymbol{\overline{\mathcal{P}}}^{\text{Mix.2}}_{\text{2D}}, \boldsymbol{\mathcal{M}}_{\text{PL}_{\text{trg}\rightarrow\text{src}}}^{B})$

	$L^{\text{trg}\rightarrow\text{src}}_{\text{Mix.3D}} = \text{CrossEntropy}(\boldsymbol{\overline{\mathcal{P}}}^{\text{Mix.2}}_{\text{3D}}, \boldsymbol{\mathcal{M}}_{\text{PL}_{\text{trg}\rightarrow\text{src}}}^{B})$

	$L^{\text{src}}_{\text{2D}\rightarrow\text{3D}} = \text{KL-Divergence}(\boldsymbol{\overline{\mathcal{P'}}}_{\text{3D}_{\text{src}}}, \boldsymbol{\overline{\mathcal{P}}}_{\text{2D}_{\text{src}}})$

	$L^{\text{src}}_{\text{3D}\rightarrow\text{2D}} = \text{KL-Divergence}(\boldsymbol{\overline{\mathcal{P'}}}_{\text{2D}_{\text{src}}}, \boldsymbol{\overline{\mathcal{P}}}_{\text{3D}_{\text{src}}})$

	$L^{\text{trg}}_{\text{2D}\rightarrow\text{3D}} = \text{KL-Divergence}(\boldsymbol{\overline{\mathcal{P'}}}_{\text{3D}_{\text{trg}}}, \boldsymbol{\overline{\mathcal{P}}}_{\text{2D}_{\text{trg}}})$

	$L^{\text{trg}}_{\text{3D}\rightarrow\text{2D}} = \text{KL-Divergence}(\boldsymbol{\overline{\mathcal{P'}}}_{\text{2D}_{\text{trg}}}, \boldsymbol{\overline{\mathcal{P}}}_{\text{3D}_{\text{trg}}})$

	$L^{\text{src}\rightarrow\text{trg}}_{\text{2D}\rightarrow\text{3D}} = \text{KL-Divergence}(\boldsymbol{\overline{\mathcal{P'}}}^{\text{Mix.1}}_{\text{3D}}, \boldsymbol{\overline{\mathcal{P}}}^{\text{Mix.1}}_{\text{2D}})$

	$L^{\text{src}\rightarrow\text{trg}}_{\text{3D}\rightarrow\text{2D}} = \text{KL-Divergence}(\boldsymbol{\overline{\mathcal{P'}}}^{\text{Mix.1}}_{\text{2D}}, \boldsymbol{\overline{\mathcal{P}}}^{\text{Mix.1}}_{\text{3D}})$

	$L^{\text{trg}\rightarrow\text{src}}_{\text{2D}\rightarrow\text{3D}} = \text{KL-Divergence}(\boldsymbol{\overline{\mathcal{P'}}}^{\text{Mix.2}}_{\text{3D}}, \boldsymbol{\overline{\mathcal{P}}}^{\text{Mix.2}}_{\text{2D}})$

	$L^{\text{trg}\rightarrow\text{src}}_{\text{3D}\rightarrow\text{2D}} = \text{KL-Divergence}(\boldsymbol{\overline{\mathcal{P'}}}^{\text{Mix.2}}_{\text{2D}}, \boldsymbol{\overline{\mathcal{P}}}^{\text{Mix.2}}_{\text{3D}})$

	Backward($L^{\text{src}}_{\text{2D}}$), Backward($L^{\text{src}}_{\text{3D}}$)\\
	Backward($L^{\text{trg}}_{\text{2D}}$), Backward($L^{\text{trg}}_{\text{3D}}$)\\
	Backward($\lambda_{\text{xm.src}}L^{\text{src}}_{\text{2D}\rightarrow\text{3D}}$), Backward($\lambda_{\text{xm.src}}L^{\text{src}}_{\text{3D}\rightarrow\text{2D}}$)\\
	Backward($\lambda_{\text{xm.trg}}L^{\text{trg}}_{\text{2D}\rightarrow\text{3D}}$), Backward($\lambda_{\text{xm.trg}}L^{\text{trg}}_{\text{3D}\rightarrow\text{2D}}$)\\
	
	Backward($L^{\text{src}\rightarrow\text{trg}}_{\text{Mix.2D}}$), Backward($L^{\text{src}\rightarrow\text{trg}}_{\text{Mix.3D}}$)\\
	Backward($L^{\text{trg}\rightarrow\text{src}}_{\text{Mix.2D}}$), Backward($L^{\text{trg}\rightarrow\text{src}}_{\text{Mix.3D}}$)\\
	
	Backward($\lambda_{\text{xm.m}}L^{\text{src}\rightarrow\text{trg}}_{\text{2D}\rightarrow\text{3D}}$), Backward($\lambda_{\text{xm.m}}L^{\text{src}\rightarrow\text{trg}}_{\text{3D}\rightarrow\text{2D}}$)\\
	Backward($\lambda_{\text{xm.m}}L^{\text{trg}\rightarrow\text{src}}_{\text{2D}\rightarrow\text{3D}}$), Backward($\lambda_{\text{xm.m}}L^{\text{trg}\rightarrow\text{src}}_{\text{3D}\rightarrow\text{2D}}$)\\
	Update($\boldsymbol{\mathcal{G}_{\text{2D}}}$), Update($\boldsymbol{\mathcal{G}_{\text{3D}}}$)
	
}

	\caption{Training neural networks with VFM-PL and FrustrumMixing.}
        \label{Algo-Training}
\end{algorithm}

\section{Dataset Splits and Class Mappings}
\label{apnd:dataset}
In this section, we will provide detailed information regarding the splits and scenes of each dataset. The class mapping of SEEM will also be presented.

\subsection{A2D2/SemanticKITTI Scenario}
\subsubsection{Datasets}\label{subsec:a2d2-skitti}
The A2D2~\cite{Geyer2020a2d2} dataset contains a total of 28637 frames from 20 drivers' driving footage. The point clouds were collected from three 16-layer LiDARs. The semantic label was provided in 2D images for 38 classes and the label of point clouds was generated by projecting the point cloud into an annotated point image.

\clearpage

The SemanticKITTI~\cite{behley2019semantickitti} dataset provides labeled data collected from a large-angle front camera and a 64-layer LiDAR. Following \cite{jaritz2022cross}, we use scenes {0,1,2,3,4,5,6,9,10} as the train set,  scene 7 and 8 as validation and test set respectively.

In the A2D2/SemanticKITTI domain adaptation scenario, frames from A2D2 are treated as the source domain while SemanticKITTI provides target data. Detailed data splits for this scenario are given as follows:
\begin{itemize}
	\item Source $\boldsymbol{\mathcal{S}}_{\text{Train}}$: 27695 frames
	\item Target $\boldsymbol{\mathcal{T}}_{\text{Train}}$:  18092 frames
	\item Target $\boldsymbol{\mathcal{T}}_{\text{Validation}}$: 1101 frames
	\item Target $\boldsymbol{\mathcal{T}}_{\text{Test}}$: 4071 frames
\end{itemize}

\subsubsection{Class Mapping}
By merging shared classes from both datasets, we identify 10 semantic labels: car, truck, bike, person, road, parking, sidewalk, building, nature and other objects. There are 133 semantic classes defined in SEEM outputs and we define a mapping between those labels (see Table~\ref{tab:cmap-a2d2}). Note that the semantic class `parking' is absent from SEEM classes, hence it introduces noise while training neural networks and reduces the overall mIoU during the test.

\begin{table}[t]
	\centering
	\begin{tabular}{@{}l|l@{}}
		\toprule
		Semantic Label                 & SEEM Class                                   \\ \midrule
		car                            & car, bus                                     \\ \midrule
		truck                          & truck                                        \\ \midrule
		bike                           & bicycle, motocycle                           \\ \midrule
		person                         & person                                       \\ \midrule
		road                           & road                                         \\ \midrule
		parking                        & N/A                                          \\ \midrule
		sidewalk                       & pavement-merged                              \\ \midrule
		building                       & house, building-other-merged                 \\ \midrule
		\multirow{3}{*}{nature}        & flower, fruit, tree-merged,\\
		&  mountain-merged, grass-merged,      \\ 
		& dirt-merged, rock-merged       \\ \midrule
		\multirow{2}{*}{other-objects} & trafficlight, stopsign, parkingmeter,   \\
		& bird, cat, dog, horse, sheep, cow, light          \\ \bottomrule
	\end{tabular}
	\caption{Class Mapping of A2D2/SemanticKITTI}
	\label{tab:cmap-a2d2}
\end{table}

\subsection{VirtualKITTI/SemanticKITTI Scenario}
\subsubsection{Datasets}
The VirtualKITTI~\cite{gaidon2016virtual} dataset comprises of 5 driving scenes generated through the Unity game engine by real-to-virtual cloning of scenes 1, 2, 6, 18 and 20 from the real KITTI dataset. The bounding box annotations from the real dataset were used to allocate cars in the virtual world. Unlike the real-world counterpart, VirtualKITTI does not simulate LiDAR, instead, it provides a dense depth map, alongside semantic, instance and flow ground truth. The 5 scenes contain a total of 2126 frames and each frame is rendered in 6 weather/lighting conditions (clone, morning, sunset, overcast, fog, rain). The whole VirtualKITTI dataset is utilized as source domain and weather/lighting conditions will be randomly sampled during training.

For details on SemanticKITTI, please refer to Sec.~\ref{subsec:a2d2-skitti}. We use the same split as A2D2/SemanticKITTI scenario, i.e. scenes {0, 1, 2, 3, 4, 5, 6, 9, 10} for training, 7 for validation and 8 for testing. Detailed splits are given as follows:
\begin{itemize}
	\item Source $\boldsymbol{\mathcal{S}}_{\text{Train}}$: 2126 frames
	\item Target $\boldsymbol{\mathcal{T}}_{\text{Train}}$:  18092 frames
	\item Target $\boldsymbol{\mathcal{T}}_{\text{Validation}}$: 1101 frames
	\item Target $\boldsymbol{\mathcal{T}}_{\text{Test}}$: 4071 frames
\end{itemize}

\subsubsection{Class Mapping}
We select 6 shared semantic labels between the two datasets. They are vegetation$\_$terrain, building, road, object, truck, car. Then we map SEEM generated classes to these semantic labels (see Table~\ref{tab:cmap-vk}).

\begin{table}[t]
	\centering
	\begin{tabular}{@{}l|l@{}}
		\toprule
		Semantic Label                       & SEEM Class                           \\ \midrule
		\multirow{2}{*}{vegetation\_terrain} & flower, fruit, tree-merged,          \\
		& grass-merged, dirt-merged            \\ \midrule
		building                             & house, building-other-merged         \\ \midrule
		road                                 & road                                 \\ \midrule
		\multirow{2}{*}{object}              & trafficlight, firehydrant, stopsign, \\
		& parkingmeter, bench, light           \\ \midrule
		truck                                & truck                                \\ \midrule
		car                                  & car                                  \\ \bottomrule
	\end{tabular}
	\caption{Class Mapping of VirtualKITTI/SemanticKITTI}
	\label{tab:cmap-vk}
\end{table}

\subsection{nuScenesLidarseg Scenarios}
The nuScenes dataset \cite{caesar2020nuscenes} contains a total of 1000 driving scenes, each of 20 seconds, which corresponds to 40000 annotated key frames taken at the frequency of 2Hz. The scenes are split into 28130 keyframes for training set, 6019 for validation set and the hidden test set. nuScenes-Lidarseg provides the point-wise label for 3D point clouds. The images taken by front camera are used in conjunction with point clouds and we adopt two scenarios for domain adaptation.
\subsubsection{USA/Singapore}
In this scenario, we use location to differentiate the source and the target domain, i.e. USA and Singapore respectively. Details of selected keyframe splits are given as follows:
\begin{itemize}
	\item Source $\boldsymbol{\mathcal{S}}_{\text{Train}}$: 15695 frames
	\item Target $\boldsymbol{\mathcal{T}}_{\text{Train}}$:  9665 frames
	\item Target $\boldsymbol{\mathcal{T}}_{\text{Validation}}$: 2770 frames
	\item Target $\boldsymbol{\mathcal{T}}_{\text{Test}}$: 2929 frames
\end{itemize}

\subsubsection{Day/Night}
In this scenario, we use the time period to separate the source and the target domain, i.e. Day and Night respectively. Details of selected keyframe splits are given as follows:
\begin{itemize}
	\item Source $\boldsymbol{\mathcal{S}}_{\text{Train}}$: 24745 frames
	\item Target $\boldsymbol{\mathcal{T}}_{\text{Train}}$:  2799 frames
	\item Target $\boldsymbol{\mathcal{T}}_{\text{Validation}}$: 606 frames
	\item Target $\boldsymbol{\mathcal{T}}_{\text{Test}}$: 602 frames
\end{itemize}

\begin{table}[t]
        \centering
	\begin{tabular}{@{}l|l@{}}
		\toprule
		Semantic Label           & SEEM Class                               \\ \midrule
		\multirow{2}{*}{vehicle} & bicycle, car, motorcycle, bus, train,    \\
		& truck                                    \\ \midrule
		building                 & house, building-other-merged             \\ \midrule
		driveable\_surface       & road                                     \\ \midrule
		sidewalk                 & playingfield, pavement-merged            \\ \midrule
		vegetation               & grass-merged, dirt-merged                \\ \midrule
		\multirow{8}{*}{manmade} & trafficlight, firehydrant, stopsign,     \\
		& parkingmeter, chair, floor-wood,         \\
		& tent, towel, bench, house, light,        \\
		& wall-stone, wall-tile, wall-wood,        \\
		& water-other, wall-other-merged,          \\
		& wall-bricks, window-blind, \\
		& fence-merged,window-other, \\
		& building-other-merged      \\ \bottomrule
	\end{tabular}
	\caption{Class Mapping of nuScenes-Lidarseg}
	\label{tab:cmap-lidarseg}
\end{table}

\subsubsection{Class Mapping}
As the number of points in the target split (e.g. for night) can
be very small for some classes, we group the classes bicycle, bus, car, construction vehicle, motorcycle, trailer, truck under the semantic label \underline{vehicle}. The barrier, pedestrian, traffic cone and other flat classes are ignored. Hence, the final 6 class labels of nuScenes-Lidarseg dataset are identified as vehicle, driveable$\_$surface, sidewalk, terrain, manmade and vegetation. We then map SEEM-generated classes to these labels (see Table~\ref{tab:cmap-lidarseg}).

\section{Additional Results on VFMSeg}
\label{apnd:additionalExp}

\subsection{Effects of the Sampled Mask Quantity}
In this section, we further investigate the effects of different quantities of sampled SAM masks. We set up three proportions to compare: large (4/5), medium (3/5) and small (1/3). The experimental results are presented in Table~\ref{tab:sam-ratio}. The medium proportion consistently outperforms the other two, hence we utilize this setting throughout the comparison with other methods in Table~\ref{tab:mainResults}.

\begin{table}[t]
\centering
\begin{tabular}{@{}ccccc@{}}
\toprule
\multicolumn{5}{c}{nuScenesLidarseg:Day/Night}                             \\ \midrule
\multicolumn{1}{c|}{No.} & \multicolumn{1}{c|}{Prop.} & 2D   & 3D   & Avg. \\ \midrule
\multicolumn{1}{c|}{\#1} & \multicolumn{1}{c|}{4/5}   & 60.0 & 70.2 & 66.0 \\ \midrule
\multicolumn{1}{c|}{\#2} & \multicolumn{1}{c|}{3/5}   & 60.6 & 70.5 & 66.5 \\ \midrule
\multicolumn{1}{c|}{\#3} & \multicolumn{1}{c|}{1/3}   & 60.6 & 70.0 & 65.9 \\ \bottomrule
\end{tabular}
\caption{Effects of different sample proportions of SAM masks. Row \#1 to Row \#3 shows the testing results of FrustrumMixing utilizing 4/5, 3/5 and 1/3 proportion of SAM segmented masks respectively.}
\label{tab:sam-ratio}
\end{table}

\begin{table}[t]
\centering
\begin{tabular}{@{}ccccc@{}}
\toprule
\multicolumn{5}{c}{VirtualKITTI/SemanticKITTI}                             \\ \midrule
\multicolumn{1}{c|}{FrustrumMixing Masks} & \multicolumn{1}{c|}{Prop.} & 2D   & 3D   & Avg. \\ \midrule
\multicolumn{1}{c|}{SEEM} & \multicolumn{1}{c|}{2/5}   & 56.5 & 50.3 & 59.6 \\ \midrule
\multicolumn{1}{c|}{SAM} & \multicolumn{1}{c|}{3/5}   & 57.2 & 52.0 & 61.0 \\ \bottomrule

\end{tabular}
\caption{Effects of substitute SAM with SEEM for generating masks to guide mixing of image and point clouds. We use the proportion of sampled SAM and SEEM masks that are estimated to cover similar overall areas.}
\label{tab:sam-seem}
\end{table}

\subsection{Additional Comparison of FrustrumMixing}
Since we are using 2D masks to generate mixed samples across different domains and modalities, there are two potential questions that need to be answered:
(1) Could we just use SEEM to guide the generation of mixed data? (2) Are there any better alternatives for SAM to perform FrustrumMixing? 

In order to answer the first question, we conduct an additional experiment on the effects of SEEM-guided mixing. We follow the identical procedure of generating mixed data with SAM and considering SEEM-generated masks are coarser than SEEM, the proportion of sampled SAM and SEEM masks are estimated beforehand so that they cover similar overall areas of input images. The results in Table~\ref{tab:sam-seem} shows that when substituting SAM with SEEM, the performances of all modalities have dropped a considerable margin which validates our choice of SAM for assisting the FrustrumMixing.

To answer the second question, we substitute CutMix~\cite{yun2019cutmix} for SAM in FrustrumMixing. The experimental results are recorded in Table~\ref{tab:sam-cutmix}. We utilize two different proportion settings in this experiment. The masked and cut out areas are estimated to contain similar labeled points/pixels, i.e. carrying a similar amount of supervised signal from one domain into the other, for generating mixed samples. We find that SAM-guided FrustrumMixing is superior in performance in both proportion settings.

\begin{table}[t]
\begin{tabular}{@{}ccccccc@{}}
\toprule
\multicolumn{7}{c}{nuScenesLidarseg:Day/Night}  \\ \midrule
\multirow{2}{*}{}     & \multicolumn{1}{c}{\multirow{2}{*}{}} & \multicolumn{2}{c}{FrustrumMixing} & \multicolumn{1}{c}{\multirow{2}{*}{}} & \multicolumn{1}{c}{\multirow{2}{*}{}} & \multicolumn{1}{c}{\multirow{2}{*}{}} \\
\cmidrule(lr){3-4}
 \multicolumn{1}{c}{No.}  & \multicolumn{1}{c}{Prop.}      & SAM    & CutMix  & \multicolumn{1}{c}{2D}    & \multicolumn{1}{c}{3D}       & \multicolumn{1}{c}{Avg.} \\ \midrule
\multicolumn{1}{c|}{\#1} & \multicolumn{1}{c|}{3/5}                   & \ding{52}      & \multicolumn{1}{c|}{}     & 60.6                                    & 70.5                                    & 66.5                                      \\ \midrule
\multicolumn{1}{c|}{\#2} & \multicolumn{1}{c|}{1/3}                   &        & \multicolumn{1}{c|}{\ding{52}}    & 59.9                                    & 69.3                                    & 65.1                                      \\ \midrule
\multicolumn{1}{c|}{\#3} & \multicolumn{1}{c|}{1/3}                   & \ding{52}      & \multicolumn{1}{c|}{}     & 60.6                                    & 69.9                                    & 65.9                                      \\ \midrule
\multicolumn{1}{c|}{\#4} & \multicolumn{1}{c|}{1/5}                   &        & \multicolumn{1}{c|}{\ding{52}}    & 59.4                                    & 69.1                                    & 63.9                                      \\ \bottomrule
\end{tabular}
\caption{Effects of substitute SAM guided mixing with CutMix in FrustrumMixing. Row \#1 and Row \#3 represents the proposed FrustrumMixing with SAM segmented masks. Row \#2 and Row \#4 represents the alternative for SAM masks, i.e. using CutMix~\cite{yun2019cutmix} to generate mixed 2D image and 3D point clouds. Note that the proportion for Row \#1 and Row \#3 stands for the ratio of sampled quantity to all SAM generated masks, while for CutMix the proportion means the ratio of cut out area to the full image size. The proportion figures were estimated to contain similar quantity of semantic labels for Row \#1/Row \#3 and Row \#2/Row \#4.}
\label{tab:sam-cutmix}
\end{table}

\subsection{Comparison of Alternative Data Augmentation}
To further demonstrate the effectiveness of the proposed FrustrumMixing, we first investigate data augmentation in an uni-modal fashion. Specifically, we perform SAM-guided 2D mixed samples, CutMix-assisted 2D augmentation and Mix3D~\cite{nekrasov2021mix3d} enhanced point cloud branch. Then, we simultaneously utilize CutMix and Mix3D for image and point cloud augmentation respectively. The experimental results are shown in Table~\ref{tab:data-aug}.

\begin{table}[t]\small
\centering
\begin{tabular}{@{}cccccccc@{}}
\toprule
\multicolumn{8}{c}{nuScenesLidarseg:Day/Night}                                                                         \\ \midrule
\multicolumn{1}{c|}{No.} & \multicolumn{1}{c|}{Prop.} & SAM & CutMix & \multicolumn{1}{c|}{Mix3D} & 2D   & 3D   & Avg. \\ \midrule
\multicolumn{1}{c|}{\#1} & \multicolumn{1}{c|}{3/5}   & \ding{52}   &        & \multicolumn{1}{c|}{}      & 60.3 & 69.1 & 65.8 \\ \midrule
\multicolumn{1}{c|}{\#2} & \multicolumn{1}{c|}{1/3}   & \ding{52}    &        & \multicolumn{1}{c|}{}      & 60.5 & 69.2 & 65.9 \\ \midrule
\multicolumn{1}{c|}{\#3} & \multicolumn{1}{c|}{1/3}   &     & \ding{52}       & \multicolumn{1}{c|}{}      & 59.7 & 68.9 & 64.2 \\ \midrule
\multicolumn{1}{c|}{\#4} & \multicolumn{1}{c|}{1/5}   &     & \ding{52}       & \multicolumn{1}{c|}{}      & 59.6 & 68.5 & 64.1 \\ \midrule
\multicolumn{1}{c|}{\#5} & \multicolumn{1}{c|}{N/A}   &     &        & \multicolumn{1}{c|}{\ding{52} }     & 59.3 & 69.1 & 65.5 \\ \midrule
\multicolumn{1}{c|}{\#6} & \multicolumn{1}{c|}{1/3}   &     & \ding{52}       & \multicolumn{1}{c|}{\ding{52} }     & 60.1 & 69.3 & 65.2 \\ \midrule
\multicolumn{5}{l}{Ours  (SAM Prop. = 3/5)}                                                        & \textbf{60.6} & \textbf{70.5} & \textbf{66.5} \\ \bottomrule
\end{tabular}
\caption{Effects of different data augmentation methods. Row \#1 to Row \#4 shows the results of 2D data augmentation via SAM and CutMix~\cite{yun2019cutmix}. Row \#5 represents 3D data augmentation that utilizes Mix3D~\cite{nekrasov2021mix3d}. Row \#6 represents the method with CutMix as 2D branch data augmentation and Mix3D for 3D branch data augmentation. }
\label{tab:data-aug}
\end{table}


\section{VFMs Generated Masks}
\label{apnd:vfmcomparison}
\begin{figure*}[ht]
	\centering
	\includegraphics[width=1\linewidth]{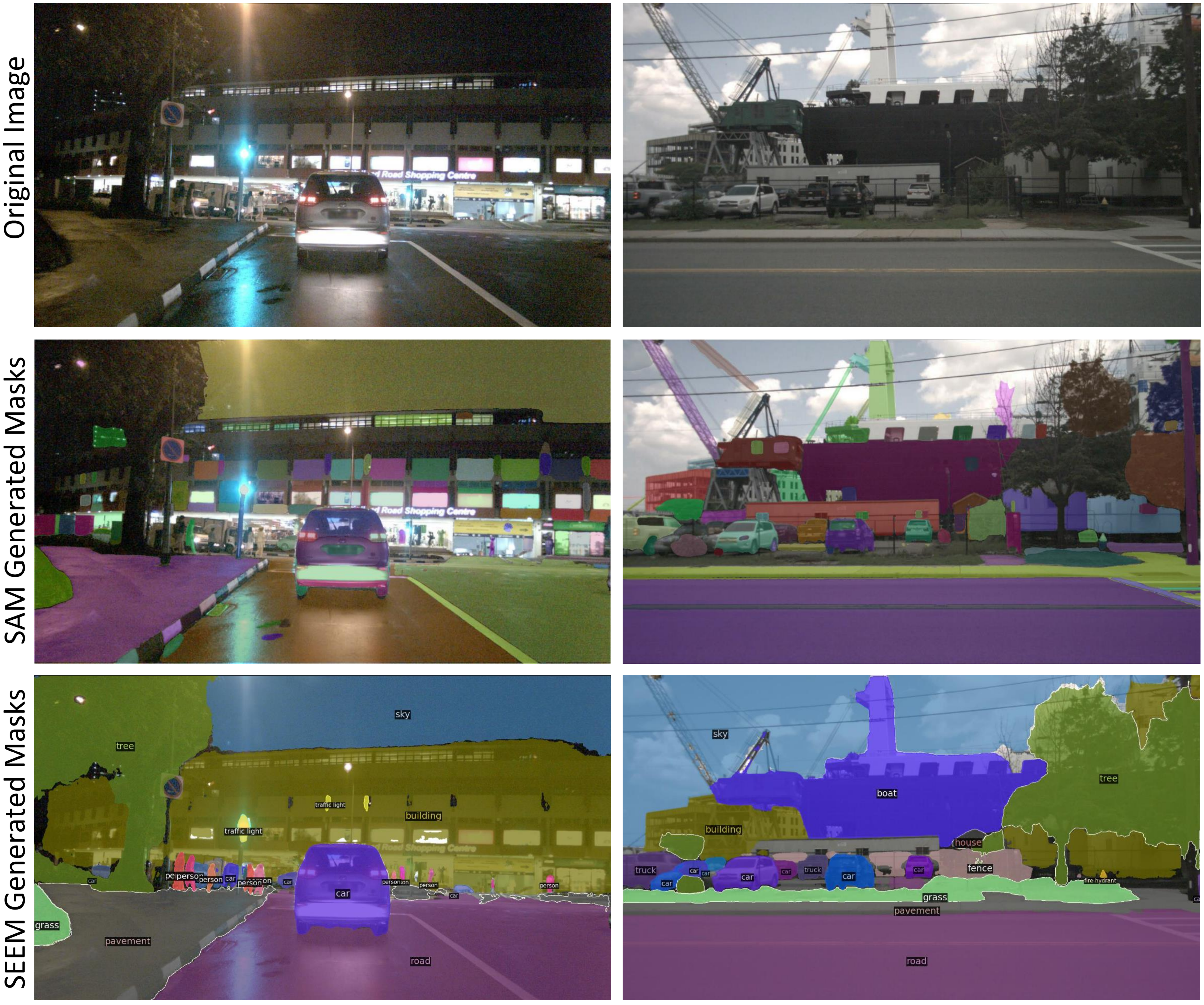}
	\caption{\textbf{Comparison between SAM-generated masks and SEEM-generated masks.} We use the official code to produce these results. Both images are sampled from the nuScenesLidarseg dataset. SAM generated masks are fine-grained and SEEM generated masks provide precise semantic labels for objects.}
	\label{fig:apnd-sam-seem}
\end{figure*}

As shown in Fig.~\ref{fig:apnd-sam-seem}, we utilize two images sampled from nuScenes-Lidarseg dataset to illustrate our incentives for leveraging both SAM and SEEM. From these samples, we can instantly recognize their remarkable capability for precise zero-shot 2D segmentation.  Specifically, SAM-generated masks are fine-grained but label-free. SEEM-generated masks could provide labels with holistic semantics for all sorts of objects. In that regard, we propose VFMSeg in this paper for enhancing the cross-modal UDA framework for 3D semantic segmentation.

\section{Additional Visualization Results}
\label{apnd:viz}
In this section, we demonstrate more qualitative results of our proposed VFMSeg to illustrate the effectiveness of our cross-modal UDA framework. The corresponding qualitative results from comparison methods will also be provided. 

\subsection{Scenario: A2D2/SemanticKITTI}
\begin{figure*}[ht]
	\centering
	\includegraphics[width=1\linewidth]{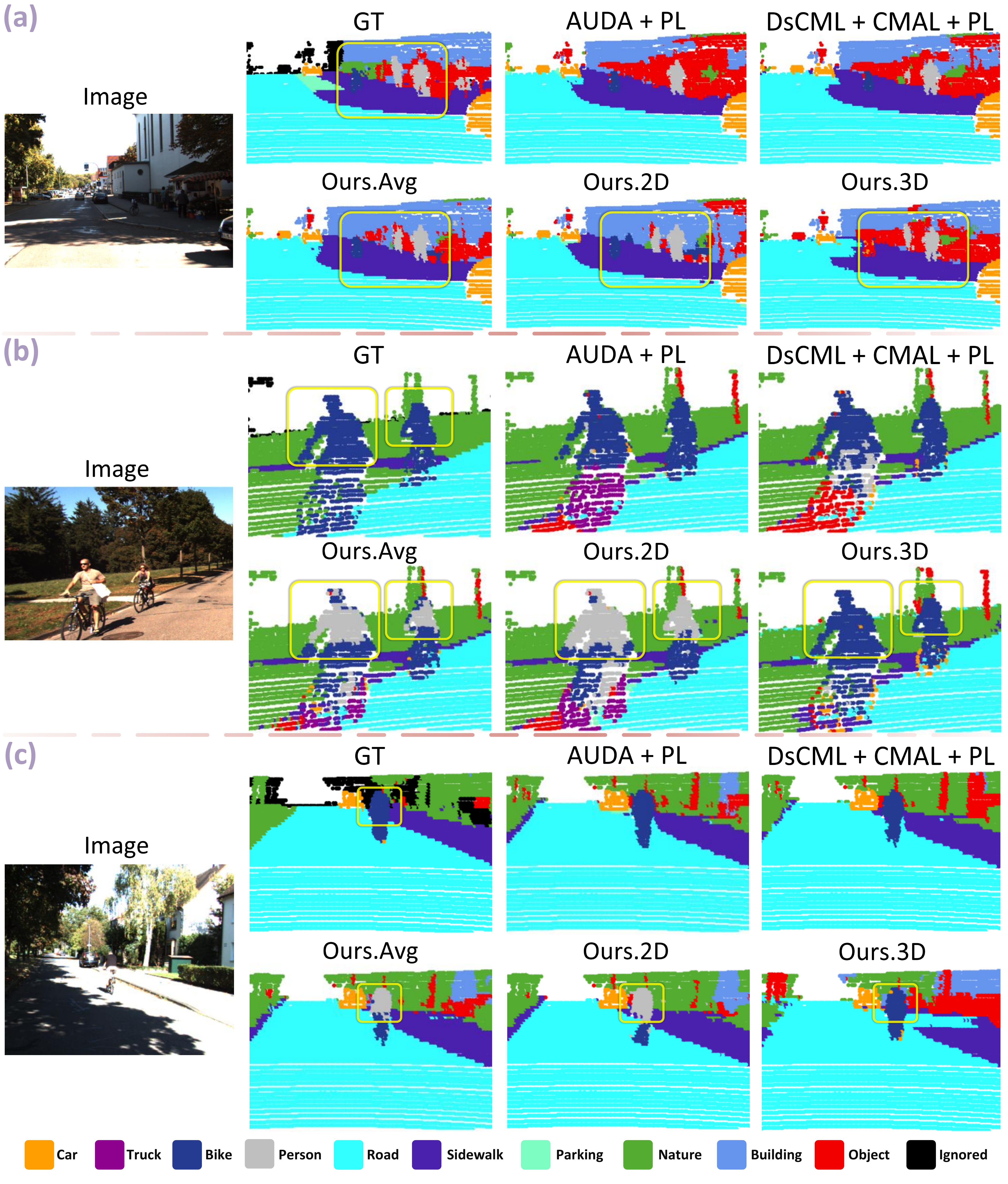}
	\caption{\textbf{Additional qualitative results for A2D2/SemanticKITTI scenario.} (a) Our method correctly identifies two people near the building; (b) The ground truth marks two people riding bikes under the label `Bike'. But our method could provide fine-grained semantics by identifying the human part from those bikes; (c) Similarly, our method provides more precise semantic labels.}
	\label{fig:apnd-a2d2}
\end{figure*}

In Fig.~\ref{fig:apnd-a2d2}, our method demonstrates the capability for generating fine-grained yet precise labels, which could provide evidence of the benefits of leveraging both SAM-predicted masks and SEEM annotated labels.

\subsection{Scenario: VirtualKITTI/SemanticKITTI}
\begin{figure*}[ht]
	\centering
	\includegraphics[width=1\linewidth]{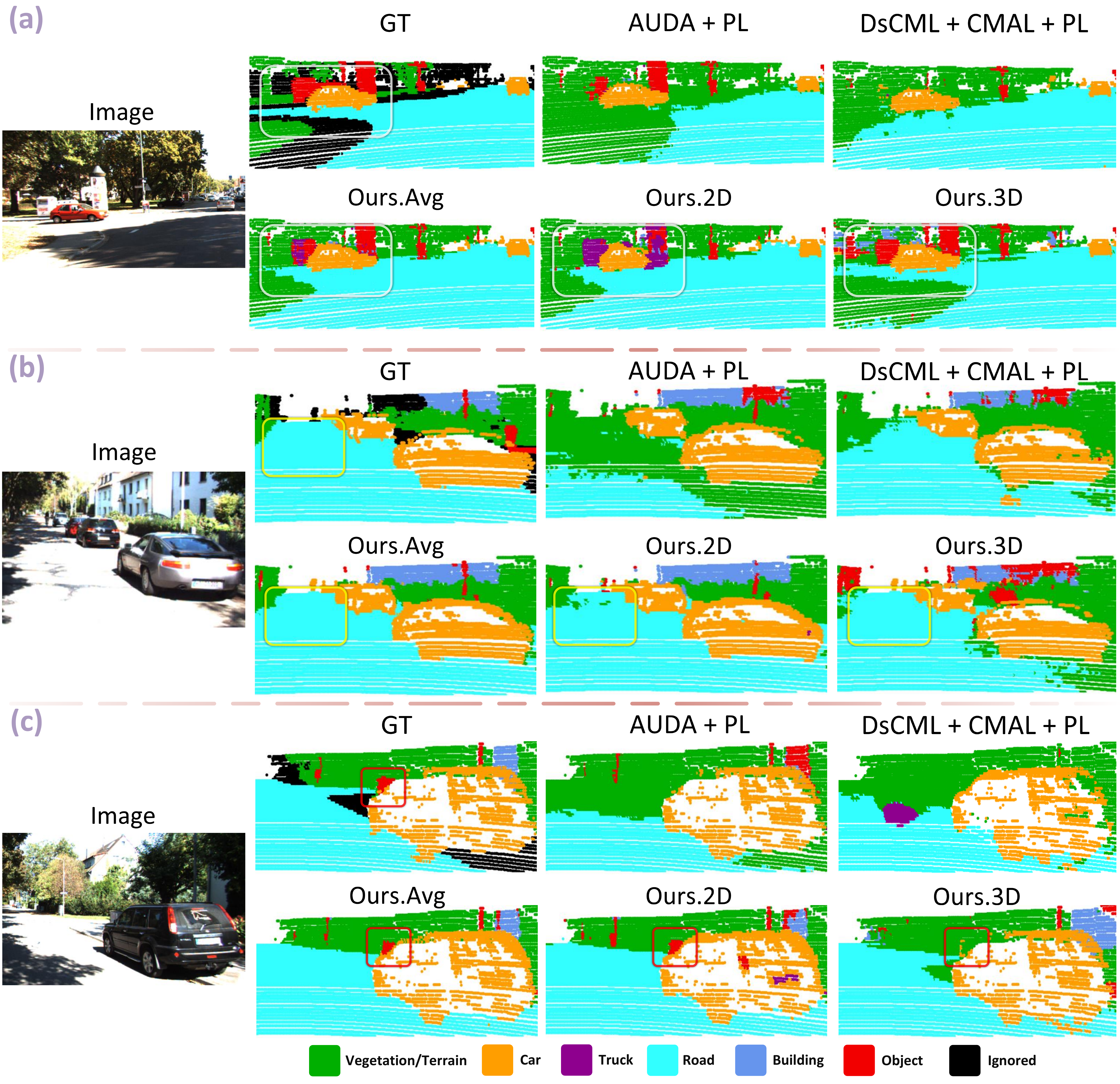}
	\caption{\textbf{Additional qualitative results for VirtualKITTI/SemanticKITTI scenario.} (a) Our method could identify the driving path on the upper left area. The 3D branch corrects the prediction of manmade objects that are wrongly labeled by 2D neural network; (b) By averaging 2D and 3D prediction, our method could accurately segment the road; (c) The 2D outputs help identify the object.}
	\label{fig:apnd-vk}
\end{figure*}

In Fig.~\ref{fig:apnd-vk}, our method could identify the road with precision and segment out objects adjacent to the vehicle. We observe that combining cross-modal predictions could improve the overall performance.

\subsection{Scenario: nuScenesLidarseg: USA/Singapore}
\begin{figure*}[ht]
	\centering
	\includegraphics[width=1\linewidth]{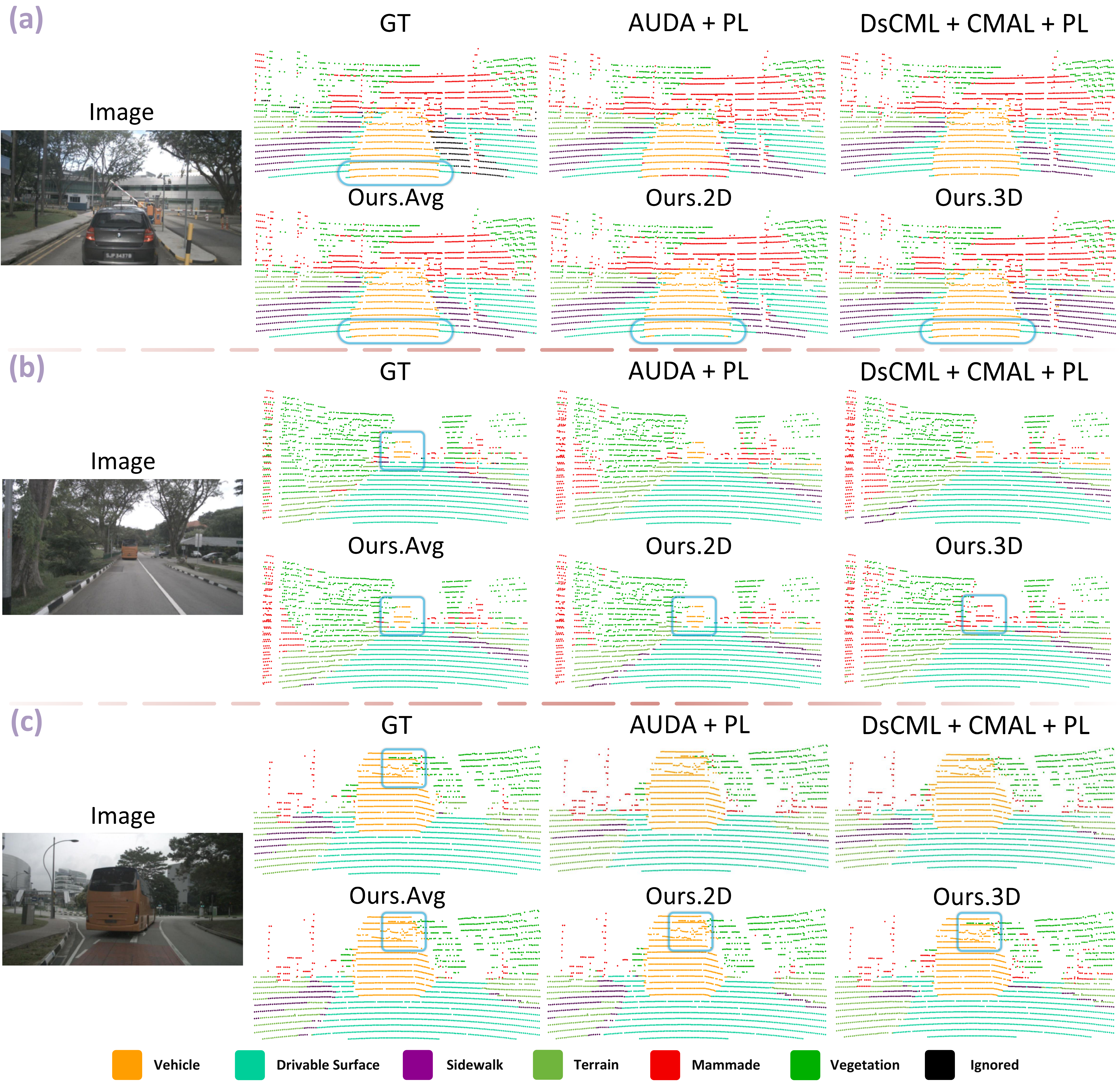}
	\caption{\textbf{Additional qualitative results for USA/Singapore scenario.} (a) Information from 3D modality promotes the overall accuracy at the bottom area of a vehicle; (b) Information from 2D modality helps identify the bus ahead; (c) The prediction from visual camera provides more accurate semantic labels at mixed areas.}
	\label{fig:apnd-usa}
\end{figure*}

In Fig.~\ref{fig:apnd-usa}(a), the learned 3D neural network corrects the prediction from the 2D branch around the bottom area of a car. From Fig.~\ref{fig:apnd-usa}(b) and Fig.~\ref{fig:apnd-usa}(c) we can see that, the 2D neural network trained under VFM-PL guidance could provide useful information about objects that are located far away and surrounded by complex background.

\subsection{Scenario: nuScenesLidarseg: Day/Night}
\begin{figure*}[ht]
	\centering
	\includegraphics[width=1\linewidth]{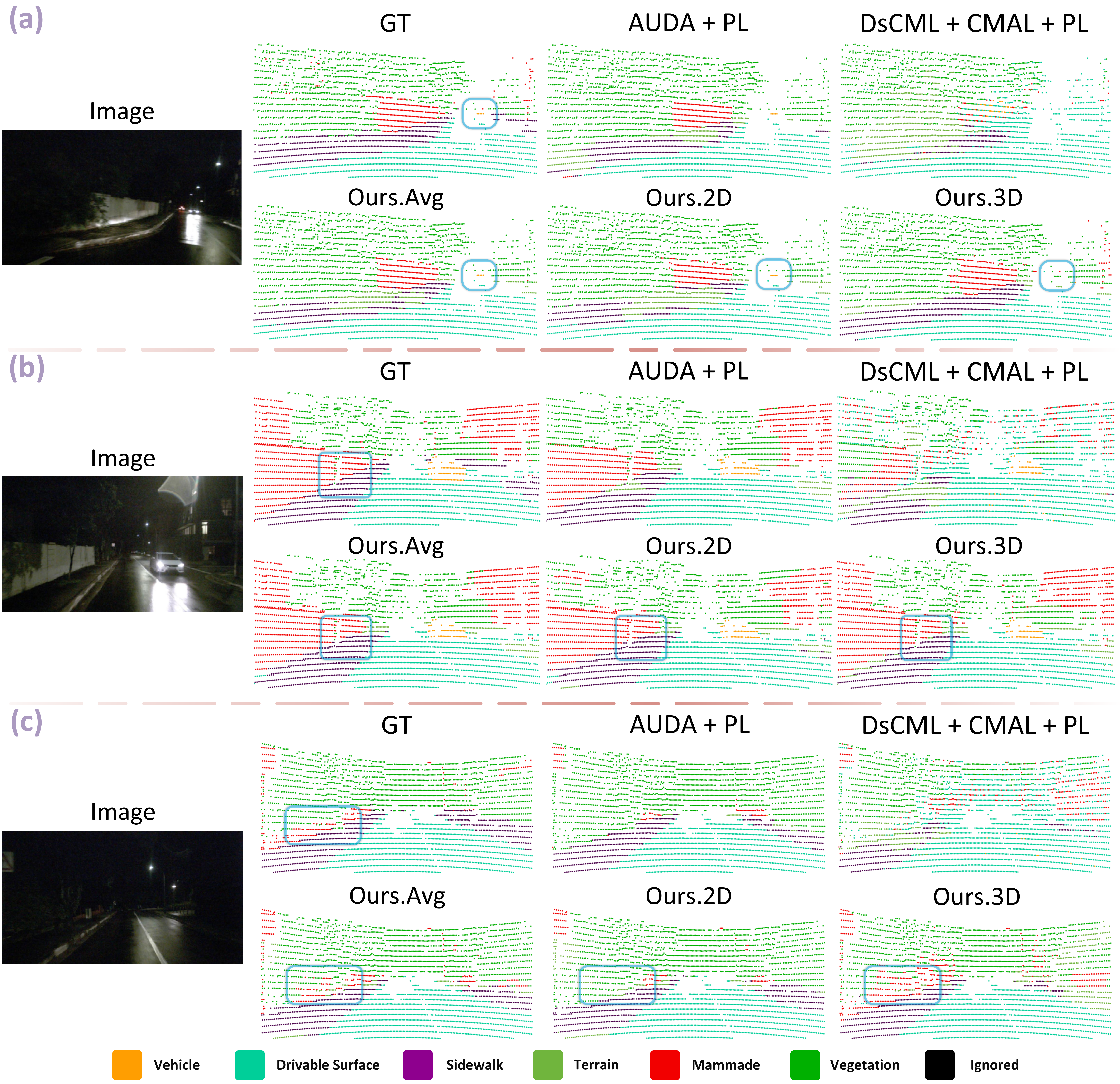}
	\caption{\textbf{Additional qualitative results for Day/Night scenario.} (a) With the help of 2D neural network, our method could identify the car ahead of the road; (b) Our method could identify the correct shape of the sidewalk; (c) 3D prediction compensates the performance of 2D neural network under restricted lighting conditions.}
	\label{fig:apnd-day}
\end{figure*}

As shown in Fig.~\ref{fig:apnd-day}, our method could mitigate the clear domain gap between day and night, and generate consistent prediction even under restricted and complex lighting conditions. By leveraging both 2D and 3D modalities and visual priors from two VFMs, the overall performance of 3D segmentation could be boosted by a significant margin.


%
%

\end{document}